\documentclass[fleqn,10pt]{wlscirep}
\usepackage[utf8]{inputenc}
\usepackage[T1]{fontenc}
\usepackage{float}
\usepackage{bbm}
\newcommand{\RW}[1]{\textcolor{black}{#1}}
\title{\RW{An Empirical Study of Machine Learning Robustness and Scalability for Imbalanced Tabular Clinical Data in Emergency and Critical Care}}

\author[1,*]{Yusuf Brima}
\author[2,3]{Marcellin Atemkeng}
\affil[1]{Computer Vision Group, Institute of Cognitive Science, Osnabrück University, Osnabrueck, D-49090, Germany}
\affil[2]{Department of Mathematics, Rhodes University, Grahamstown 6139, South Africa}
\affil[3]{National Institute for Theoretical and Computational Sciences (NITheCS), Stellenbosch 7600, South Africa}
\affil[*]{ybrima@uos.de}

\newcommand{\SOTA}{SOTA}
\newcommand{\FOneMacro}{F1-Score}

\newcommand{\MIMIC}{\textit{MIMIC-IV-ED}}

\begin{abstract}
Every year, millions of patients pass through emergency departments and intensive care units where clinicians must make life-altering decisions under time pressure and uncertainty. Advances in \RW{machine learning} is poised to offer \RW{support for clinical decision-making, including prediction of patient deterioration, triage guidance, and identification of rare but clinically critical outcomes}. Yet a persistent impediment limits its utilization in these settings: clinical data are often severely imbalanced, with critical outcomes occurring far less frequently than routine ones. This skewness can bias models toward majority classes, degrading performance. Developing models that are both robust to such imbalance and computationally efficient enough for deployment in time-sensitive environments remains an open and practically important challenge.

In this paper, we empirically studied the robustness and scalability of \RW{six model families spanning classical machine learning, deep learning, and tabular foundation models} on imbalanced tabular data from two large-scale clinical datasets (MIMIC-IV-ED and eICU). Class imbalance was quantified using three complementary metrics, and we compared tree-based methods (Decision Tree, Random Forest, XGBoost), the TabNet deep learning model, and two tabular foundation models (TabICL and TabPFN v2.6). \RW{All trainable models were evaluated under a unified experimental protocol using Bayesian hyperparameter optimization for trainable models, while foundation models were assessed in their pretrained inference regime without task-specific optimization or reweighting.} All models were assessed on predictive performance (\RW{Macro F1-score}), robustness to increasing imbalance, and computational scalability across seven clinically relevant prediction tasks.

Results differed across databases. On MIMIC-IV-ED, foundation-based models (TabPFN v2.6 and TabICL) attained the strongest average \RW{Macro F1-score} ranks, with XGBoost and other tree-based ensembles remaining competitive. On eICU, XGBoost consistently led, followed by other tree-based methods, while foundation models occupied intermediate positions. Across both datasets, TabNet exhibited the sharpest performance degradation under increasing imbalance and the highest computational costs. Training time analyses showed that classical and tree-based methods scale most favorably with dataset size, while foundation models achieved low per-task cost through their inference-based paradigm.

These findings indicate that model selection for imbalanced clinical tabular data is context-dependent: no single family dominated across both datasets and all tasks. Nonetheless, recent advances in tabular foundation models suggest a rapidly narrowing performance gap with strong classical baselines such as XGBoost, while offering a distinct computational profile characterized by low per-task adaptation cost. This efficiency–performance trade-off may become increasingly relevant for deployment in resource-constrained clinical environments. Rather than prescribing a universal solution, this work provides clinical stakeholders with an empirically grounded framework for navigating the trade-offs between predictive robustness, computational scalability, and clinical feasibility in high-stakes, time-sensitive care environments.

\vspace{1.5pt}
\end{abstract}

\begin{document}
\flushbottom
\maketitle
\thispagestyle{empty}

\section*{Introduction}

In the emergency department (ED) and intensive care unit (ICU), clinicians operate under conditions characterized by high patient turnover, unpredictable workloads, and the need for rapid decision-making in life-threatening situations~\cite{huang2010impact,carayon2014human,weigl2012association}. In such settings, even small delays can have significant consequences, motivating the development of decision support systems that provide accurate predictions with minimal computational overhead~\cite{johnson2016machine,komorowski2018artificial}. Machine learning (ML), a subfield of artificial intelligence (AI), has emerged as a promising approach for supporting clinical decision-making under these constraints~\cite{esteva2019guide,topol2019high,rajpurkar2022ai,miotto2018deep,jiang2017artificial}, with applications spanning diagnosis, prognosis, triage, and patient disposition.

However, healthcare data present several persistent challenges. In addition to \emph{high dimensionality} and \emph{heterogeneity}, a defining characteristic is \emph{class imbalance}, where clinically critical outcomes (e.g., in-hospital mortality, cardiac arrest, or septic shock) occur relatively rarely compared to more common outcomes. This imbalance can lead to models that achieve strong aggregate performance while under-performing on minority classes, which are often of greatest clinical interest~\cite{He2009,Branco2016}.

These characteristics have important implications for model selection. Deep learning (DL), a subset of ML, has demonstrated transformative performance in domains such as computer vision and natural language processing~\cite{lecun2015deep,krizhevsky2012imagenet,silver2017mastering}, and more recently, a range of architectures have been proposed for tabular data. These include attention-based models such as TabNet~\cite{Arik2021}, as well as emerging tabular foundation models that leverage large-scale pre-training and in-context learning. While these approaches show promise, their behavior on structured clinical data particularly under varying degrees of class imbalance and operational constraints remains an active area of investigation.

In parallel, classical ML methods, especially tree-based ensemble techniques such as random forests and gradient boosting (e.g., XGBoost), continue to be widely used as pragmatic tools in tabular healthcare applications due to their strong empirical performance and relatively understood stable behavior across diverse settings~\cite{Chen2016,Lundberg2020}. Prior comparisons between ML and DL approaches in healthcare have reported mixed findings: while DL models may achieve improved performance under certain conditions (e.g., large-scale datasets and extensive tuning), classical approaches often remain competitive, particularly in imbalanced or moderately sized tabular datasets~\cite{Choi2017,Katuwal2016,Luo2016}. However, these comparisons are often limited by differences in experimental design, dataset selection, or the range of models considered.

To address class imbalance, a variety of strategies have been proposed, including resampling methods such as Synthetic Minority Over-sampling TEchnique (SMOTE)~\cite{chawla2002smote}, cost-sensitive learning~\cite{elkan2001foundations}, and modified loss functions such as focal loss~\cite{lin2017focal,Buda2018}. Although these approaches have been widely studied, their comparative behavior across different tabular learning paradigms and heterogeneous clinical datasets remains insufficiently understood, particularly under realistic imbalance conditions.

\RW{Against this backdrop, this study provides a systematic empirical characterization of how class imbalance influences predictive performance and computational scalability across diverse ML paradigms for tabular clinical data in emergency and critical care settings. Using two large-scale clinical databases, MIMIC-IV-ED and eICU, we evaluate model behavior across seven clinically relevant prediction tasks spanning diagnosis, disposition, severity, and operational outcomes. Our analysis investigates representative classical ML methods, deep learning architectures, and emerging inference-based tabular foundation models under controlled variations in class imbalance and dataset scale.}

\RW{Specifically, this work makes four main contributions:}

\begin{itemize}
    \item \RW{We provide a cross-institutional empirical evaluation of class imbalance effects across multiple tabular learning paradigms using two large-scale datasets (MIMIC-IV-ED and eICU) over seven clinically relevant prediction tasks.}
    
    \item \RW{We systematically varied ML paradigms under realistic imbalance conditions, while examining the impact of multiple imbalance quantification metrics and weighting strategies on predictive degradation.}
    
    \item \RW{We analyze computational efficiency and scalability across model families, including the distinct operational characteristics of inference-only foundation models such as TabPFN and TabICL relative to conventionally trainable models.}
    
    \item \RW{We also expound on useful insights into model selection under competing constraints of imbalance robustness, computational cost, scalability, interpretability, and deployment feasibility in real-world clinical environments.}
\end{itemize}

The goal of this work is to provide an empirically grounded analysis of representative approaches under consistent experimental conditions. By doing so, we aim to (i) offer pragmatic insight into the trade-offs between robustness and computational efficiency, and (ii) support more informed model selection for imbalanced tabular clinical data.

\subsection*{Organization of the Paper}

The remainder of this paper is organized as follows. The \textit{Methods} section describes the datasets, prediction tasks, model architectures, and evaluation procedures, including approaches used to quantify and tackle class imbalance. The \textit{Results} section presents the empirical findings. The \textit{Discussion} section interprets these results in the context of robustness, scalability, and potential clinical applicability, and outlines limitations and directions for future work. Finally, the \textit{Conclusion} summarizes the main contributions of the study.

\section*{Methods}
In this section, we expound on the data utilized: their sources, characterization, preprocessing, and specific predictive tasks they tackle. Thereafter, we discuss the model architectures and their configurations used. Then, the optimization algorithm, including the training objectives, is discussed. After that, we explain both the imbalance quantification metrics and classification performance metrics. Finally, the experimental setup is stated.
\subsection*{Healthcare Prediction Tasks and Datasets}
To conduct this study, our goal was to use data that are clinically and contextually relevant to the problem in question. Therefore, the data sources were chosen because of their ideal fit for this purpose. In that regard, we describe these datasets below accordingly.
\subsubsection*{MIMIC-IV-ED}
We used the \MIMIC{} database (v2.2), which contains approximately 425,000 ED stays collected between 2011 and 2019 at Beth Israel Deaconess Medical Center in Boston, Massachusetts~\cite{johnson2021mimic}. It is hosted on the PhysioNet platform~\cite{goldberger2000physiobank} and includes detailed demographic information, triage measurements, periodic vital signs, medication administrations, and discharge diagnoses. Its rich clinical coverage makes it a suitable resource for assessing ML models under realistic conditions such as class imbalance.

We defined three clinically relevant prediction tasks using this dataset. 
First, we aimed to predict the primary diagnosis at discharge, capturing the most pressing clinical issue during the ED stay. Second, we grouped diagnoses into three-character International Classification of Diseases (ICD 9 and 10) categories to assess model performance at a higher level of disease semantic abstraction, reducing label sparsity while retaining clinically meaningful distinctions. Third, we predicted ED disposition outcomes, including admission, discharge, transfer, or death, which reflect critical operational and patient safety considerations. These tasks span different levels of clinical granularity, enabling a comprehensive assessment of model robustness across diverse prediction scenarios.

To prepare the data for model training and evaluation, we applied systematic preprocessing and feature engineering steps, including handling of missing values, normalization of continuous variables, and encoding of categorical variables. We also employed stratified sampling to ensure proportional representation of all target classes in the training, validation, and test splits. Additional details of the preprocessing workflow, feature construction, and dataset assembly are provided in Appendix~\ref{appendix:preprocessing_mimic}.

\subsubsection*{eICU Collaborative Research Database (eICU-CRD)}
To complement the single-center MIMIC-IV-ED data and assess model robustness across multiple institutions, the eICU-CRD was utilized. It contains circa 200,000 ICU stays collected from multiple hospitals across the United States~\cite{pollard2019eicu}. This multi-center dataset includes patient demographics, vital signs, laboratory measurements, clinical interventions, and outcomes, providing a broad context to evaluate the robustness of models across diverse hospital settings and patient demographics.

The prediction tasks closely mirrored those defined for MIMIC-IV-ED to enable cross-dataset comparisons of methods appropriately. These included length of stay prediction, and patient disposition, such as ICU discharge, transfer, or death, etc. Maintaining comparable prediction tasks allows direct assessment of model performance under differing data distributions, class imbalances, and institutional practices.

Preprocessing and feature engineering followed the same principles applied to the prior dataset, including handling of missing values, normalization of continuous variables, and one-hot encoding of categorical variables. Stratified sampling preserved class distributions across training, validation, and test splits. Detailed descriptions of feature extraction, dataset assembly, and preprocessing for the eICU database are provided in Appendix~\ref{appendix:preprocessing_eicu}.

\subsection*{Models Evaluated}
To assess the performance of the chosen ML algorithms, a range of methods was evaluated, spanning classical ML, \RW{deep learning, and inference-based tabular foundation models}. This enabled comparisons across interpretable tree-based methods, \RW{attention-based neural architectures, and foundation-style approaches}, with the goal of assessing robustness and scalability under class imbalance.

\subsubsection*{Traditional ML Models}
Classical ML algorithms remain widely used for structured (tabular) data due to their strong empirical performance, relative interpretability, and well-understood behavior across a range of applications. In this study, we selected several representative methods to provide a baseline for comparison with more recent approaches, particularly under conditions of class imbalance.

The first of these is the \textit{decision tree} (DT) algorithm~\cite{quinlan1986induction}, which partitions the feature space into a hierarchical structure by iteratively selecting splits that maximize reductions in impurity (e.g., Gini impurity or entropy). Each internal node corresponds to a decision rule, while leaf nodes represent class predictions. While simple and interpretable, individual trees are known to be sensitive to data perturbations and may exhibit limited generalization performance.

To address these limitations, the \textit{random forest} (RF) algorithm~\cite{breiman2001random} constructs an ensemble of decision trees, each trained on a bootstrap sample of the data and a random subset of features. Predictions are obtained via aggregation (typically majority voting), which reduces variance and generally improves robustness. Ensemble approaches such as RF are often reported to perform reliably on noisy or moderately imbalanced datasets, although their performance may still degrade under more extreme imbalance conditions.

We further incorporated the \textit{XGBoost} algorithm~\cite{chen2016xgboost}, a widely used implementation of gradient boosting. In this framework, trees are added sequentially, with each new tree trained to correct the residual errors of the current ensemble. By optimizing a regularized objective function, XGBoost can capture complex feature interactions while controlling overfitting. We used XGBoost as a representative of gradient boosting methods; related implementations such as LightGBM~\cite{ke2017lightgbm} and CatBoost~\cite{prokhorenkova2018catboost} introduce algorithmic and engineering refinements, but prior comparative studies suggest broadly similar performance characteristics at the level of model families~\cite{grinsztajn2022tree}. As the focus of this work is on comparative behavior across model classes rather than exhaustive benchmarking of individual implementations, we limit our evaluation to a single representative boosting framework.

Together, these models provide a set of established candidates for tabular prediction tasks. Their inclusion enables a structured comparison with more recent deep learning and foundation-based approaches, with particular attention to how different model families behave under varying degrees of class imbalance and dataset scale.

\subsubsection*{\SOTA{} Deep Learning Models}
We evaluated \RW{contemporary neural and foundation-based approaches} for tabular data to provide a more comprehensive and up-to-date study.

The primary neural architecture considered was \textit{TabNet}~\cite{arik2021tabnet}, an attention-based deep neural network specifically designed for tabular learning. TabNet employs a sequential attention mechanism that selectively focuses on subsets of features at each decision step, enabling the model to capture complex feature interactions while maintaining a degree of interpretability through learned feature masks.

In addition, we included \textit{TabPFN}~\cite{hollmann2022tabpfn,grinsztajn2025tabpfn}, a recently proposed tabular foundation model that leverages a transformer-based prior trained on a large distribution of synthetic datasets. \RW{TabPFN was evaluated strictly as an inference-only model in its pretrained form, without task-specific optimization, retraining, fine-tuning, or class-weighting modifications. Predictions are generated directly through a single forward pass, consistent with the intended usage paradigm of TabPFN v2.6.}

We also incorporated \textit{TabICL}, an in-context learning framework for tabular data that builds upon the foundation model paradigm~\cite{qu2025tabicl}. \RW{Similar to TabPFN, TabICL was evaluated as an inference-only approach without parameter updates or training-time imbalance mitigation strategies such as class weighting.} TabICL performs prediction by conditioning on training examples directly at inference time, enabling strong performance without explicit parameter updates.

\RW{Accordingly, imbalance mitigation strategies evaluated in this study apply only to trainable models (DT, RF, XGBoost, and TabNet), whereas TabPFN and TabICL are assessed exclusively within their pretrained inference regimes.}

These additions allow for a more balanced comparison between classical learning methods, conventional DL models, and recent tabular foundation models. At the same time, we note that differences in training paradigms and computational characteristics should be considered when interpreting results, particularly in the context of real-world clinical deployment constraints.

\RW{The implementation leverages PyTorch for TabICL, TabPFN v2.6, and TabNet, whereas Scikit-learn is used for decision trees, random forests, and XGBoost.} For final reporting, pre-set random seeds that followed experiment numbers were to ensure reproducibility across models.

\subsection*{Class Imbalance Handling Strategies} 
We first introduced the notational framework used throughout this section, which subsequently allowed us to define precisely how imbalance is quantified and addressed.

Throughout, we write scalars in italics (e.g., $n, d, K$), vectors in bold lowercase (e.g., $\mathbf{x}, \mathbf{z}$), and matrices in bold uppercase (e.g., $\mathbf{X} \in \mathbb{R}^{N \times d}$). 
Sets and spaces are denoted in calligraphic font (e.g., $\mathcal{D}, \mathcal{X}, \mathcal{Y}, \mathcal{Z}$), while functions and mappings are written in standard math operator style (e.g., $f, \sigma, \mathrm{softmax}$). 
In particular, we use $\mathbf{x}_i$ to denote an \emph{individual input vector} and $\mathbf{X}$ for the \emph{design matrix} containing all samples stacked row-wise. 
This convention ensures a clear distinction between observed data $(\mathbf{x}_i,y_i)$, latent representations $\mathbf{z}_i$, predictions $\hat{y}_i$, and the mappings that connect them.

\paragraph{Dataset and Input Space}\mbox{}\\ 
Let a dataset be denoted as $\mathcal{D} := \{(\mathbf{x}_i, y_i)\}_{i=1}^N$ where $N \in \mathbb{N}$ is the total number of samples. Each input $\mathbf{x}_i \in \mathcal{X}$ belongs to the feature space $\mathcal{X} \subseteq \mathbb{R}^d$ with $d$ the dimensionality.

\paragraph{Label Space}\mbox{}\\ 
Each label $y_i \in \mathcal{Y}$, where $\mathcal{Y} = \{1,2,\dots,K\}$ is a discrete variable representing one of $K$ possible classes. For multi-class classification, $y_i$ may be equivalently represented as a one-hot vector in $\{0,1\}^K$. We denote the vector of all labels as $\mathbf{y} = (y_1,\dots,y_N)^\top$.

\paragraph{Latent (Logit) Space}\mbox{}\\ 
The model, in a general sense, is a function $f: \mathcal{X} \to \mathcal{Z}$ mapping feature vectors to latent representations. For each input $\mathbf{x}_i$, it produces logits $\mathbf{z}_i \in \mathcal{Z}$, where $\mathcal{Z} \subseteq \mathbb{R}^K$ in the multi-class case and $\mathcal{Z} \subseteq \mathbb{R}$ for binary classification. 
We denote vectors of logits as $\mathbf{z}_i \in \mathbb{R}^K$ and the stacked matrix as $\mathbf{Z} \in \mathbb{R}^{N \times K}$. 
These logits are transformed into probabilities through activation functions:
\begin{align*}
\sigma &: \mathbb{R} \to [0,1], && \text{(sigmoid for binary tasks)}, \\
\mathrm{softmax} &: \mathbb{R}^K \to [0,1]^K, && \text{(softmax for multi-class tasks)}.
\end{align*}
The predicted label is then obtained as:
\begin{equation*}
    \hat{y}_i =  \underset{k\in K}{\arg\max}\, \hat{y}_{i,k}.
\end{equation*}
We denote the probability vector for sample $i$ as $\hat{\mathbf{y}}_i \in [0,1]^K$ and the full prediction matrix as $\hat{\mathbf{Y}} \in [0,1]^{N \times K}$.

With this formalization in place, we now describe strategies to address class imbalance during training. 
We focused on approaches that can be applied consistently across both classical ML algorithms and deep architectures. 
While alternative methods such as focal loss~\cite{lin2017focal} have been proposed specifically for neural networks to emphasize hard-to-classify examples, they are less straightforward in tree-based models or other classical algorithms. 
To ensure comparability across model families, we implemented three complementary weighting strategies derived directly from the label distribution. 

The first, \textbf{Inverse Frequency} which is widely adopted, assigns a weight to each class inversely proportional to its number of samples in that class. 
For class $k$, the weight is computed as:
\begin{equation}
w_k = \frac{N}{K \, N_k}, 
\label{eq:if}
\end{equation}
where $N$ is the total number of training samples, $K$ the number of classes, and $N_k$ the number of samples in class $k$. 
This ensures that minority classes contribute more during training. 

The second strategy, \textbf{Effective Number of Samples}~\cite{cui2019class}, accounts for the diminishing benefit of additional samples from frequent classes. 
Let $\beta \in [0,1]$ be a smoothing factor. 
The effective number of samples for class $k$ is:
\[
N_k^\text{eff} = \frac{1 - \beta^{N_k}}{1 - \beta},
\]
with the corresponding normalized weight:
\[
w_k = \frac{1}{N_k^\text{eff}} \frac{\sum_{j=1}^K N_j^\text{eff}}{K}.
\]
This reduces the dominance of majority classes while avoiding excessively large weights for rare ones. 

The third approach, \textbf{Median Frequency Balancing}, scales the weight of each class by the ratio of the median class frequency to the class’s frequency:
\[
w_k = \frac{\mathrm{median}(f_1,\dots,f_K)}{f_k}, 
\qquad f_k = \frac{N_k}{N},
\]
where $f_k$ is the relative frequency of class $k$. 
This method balances contributions without allowing rare classes to dominate excessively.

\noindent These computed weights $\{w_k\}_{k=1}^K$ are incorporated directly into the loss functions for both binary and multi-class tasks, ensuring that minority classes exert proportional influence during optimization. 
This is supposed to improve detection of rare but clinically significant outcomes while maintaining training stability.

\subsection*{Class Imbalance Quantification} 

To evaluate model robustness systematically, we filtered the composition of each dataset to create controlled levels of class imbalance. 
This was achieved by varying the minimum number of samples required for each class within that dataset. 
Lower thresholds retain rarer classes, producing training and evaluation candidate dataset with pronounced imbalance (i.e., skewed distributions), whereas higher thresholds favor more common classes, resulting in a nearly more uniform distribution. 

To assess these effects, we quantified the degree of imbalance via three complementary metrics, each capturing a property of class representation.

\subsubsection*{Coefficient of Variation of Class Frequency (CVCF)} 
The first metric, the CVCF, measures the relative variability in class \emph{frequencies}, highlighting whether some classes dominate the dataset. 

\noindent For each class $k$, the relative frequency is calculated as:
\begin{equation}
f_k = \frac{N_k}{N}.
\label{eq:cf}
\end{equation}

\noindent Given these frequencies $\{f_k\}_{k=1}^K$, the CVCF is defined as:
\begin{equation}
\begin{aligned}
\bar{f} &= \frac{1}{K} \sum_{k=1}^K f_k && \text{(mean class frequency)}, \\
\sigma_f &= \sqrt{\frac{1}{K} \sum_{k=1}^K (f_k - \bar{f})^2} && \text{(standard deviation of class frequencies)}, \\
\text{CVCF} &= \frac{\sigma_f}{\bar{f}} && \text{(coefficient of variation)}.
\end{aligned}
\end{equation}

\noindent A higher CVCF signals pronounced imbalance, with certain classes disproportionately represented, whereas a lower CVCF reflects more uniform class distributions.

\subsubsection*{Imbalance Ratio (IR)} 
Complementing the CVCF, the IR captures the disparity between the most and least represented classes. 
Let $\{N_k\}_{k=1}^K$ denote class counts:
\begin{equation}
\text{IR} = \frac{\max_k N_k}{\min_k N_k}, \quad \min_k N_k > 0.
\end{equation}
An IR of 1 indicates perfectly balanced classes, while higher values correspond to increasingly skewed distributions. 
Unlike CVCF, which accounts for all class frequencies, IR focuses specifically on the extremes of the distribution. 

\subsubsection*{Normalized Entropy of Class Distribution (NECD)} 
While CVCF captures variability across all classes and IR emphasizes extremes, both are inherently scale-free statistics: CVCF is a ratio of dispersion to mean, and IR is a ratio of maximum to minimum class counts. 
Their values are directly comparable across problems with different numbers of classes. 
Entropy provides a complementary perspective by quantifying the uncertainty of predicting a random class label, reaching its maximum under a uniform distribution and decreasing as the distribution becomes skewed. 
Unlike CVCF and IR, however, the raw value of entropy depends on the number of classes $K$, which makes direct comparisons across tasks misleading. 
To address this, we normalize entropy by its maximum possible value, ensuring that the measure consistently reflects class balance irrespective of $K$. 

Using the relative frequencies $\{f_k\}_{k=1}^K$ defined in Equation~\ref{eq:cf}, the Shannon entropy is
\begin{equation}
H = - \sum_{k=1}^K f_k \log f_k,
\end{equation}
with $f_k \log f_k = 0$ when $f_k = 0$. 
The maximum entropy is
\begin{equation}
H_{\text{max}} = \log(K),
\end{equation}
corresponding to a perfectly uniform distribution. 
The normalized entropy is then
\begin{equation}
\text{NECD} = \frac{H}{H_{\text{max}}} = \frac{-\sum_{k=1}^K f_k \log f_k}{\log(K)}.
\end{equation}
NECD ranges from 0 (complete imbalance) to 1 (perfect balance), with intermediate values reflecting partial uniformity. 

Together with CVCF and IR, it provides a complementary measure for generating datasets with controlled imbalance and analyzing their impact on model performance.

\subsubsection*{Model Predictive Performance Evaluation} 

We evaluated the predictive performance of all models using two complementary metrics: overall accuracy and the Macro F1 score. Accuracy measures the fraction of correct predictions across all samples as shown in equation~\ref{eq:accuracy}, providing a straightforward assessment of overall model correctness. However, in datasets with class imbalance, accuracy can give a distorted view of performance because \textit{it can be dominated by the majority classes}, masking poor performance on clinically important minority classes. 

For a dataset with $N$ samples, we denote the ground-truth labels as $y_i$ and the model's predicted class as $\hat{y}_i$. Overall accuracy is computed as:
\begin{equation}
\text{Accuracy} = \frac{1}{N} \sum_{i=1}^{N} \mathbbm{1}\{y_i = \hat{y}_i\}, 
\label{eq:accuracy}
\end{equation}
where $\mathbbm{1}\{\cdot\}$ is the indicator function, equal to 1 if the condition inside is true and 0 otherwise. 

The F1 score on the other hand provides a balanced measure of precision and recall for each class. For a given class $k \in \{1,\dots,K\}$, we define:
\begin{equation*}
\text{Precision}_k = \frac{\text{TP}_k}{\text{TP}_k + \text{FP}_k}, 
\qquad 
\text{Recall}_k = \frac{\text{TP}_k}{\text{TP}_k + \text{FN}_k}, 
\end{equation*}
where $\text{TP}_k$, $\text{FP}_k$, and $\text{FN}_k$ denote the number of true positives, false positives, and false negatives for class $k$, respectively. 

The F1 score for class $k$ is then:
\begin{equation}
\text{F1}_k = \frac{2 \cdot \text{Precision}_k \cdot \text{Recall}_k}{\text{Precision}_k + \text{Recall}_k}.
\label{eq:f1_score}
\end{equation}
For multi-class tasks, we report the (weighted) F1, defined as the mean of $\text{F1}_k$ across all $K$ classes:
\[
\text{F1} = \frac{1}{K}\sum_{k=1}^K \text{F1}_k.
\]

By reporting both metrics, our goal is to ensure a more comprehensive and reliable assessment of predictive performance, capturing both the overall correctness and the model's ability to correctly identify minority classes. 

In this formulation, $y_i$ comes directly from the dataset, and $\hat{y}_i$ is obtained from the model outputs. For binary classification, the model produces a single logit $z_i \in \mathbb{R}$, which is transformed into a probability through the sigmoid function
\[
\hat{y}_i = \sigma(z_i) = \frac{1}{1 + e^{-z_i}}.
\]
For multi-class classification, the model outputs a logit vector $\mathbf{z}_i \in \mathbb{R}^K$, which is converted to a probability distribution by the softmax function:
\[
\mathrm{softmax}(z_i)_k = \frac{e^{z_{i,k}}}{\sum_{j=1}^K e^{z_{i,j}}}, 
\qquad k = 1,\dots,K.
\]
The predicted label $\hat{y}_i$ is then obtained by selecting the most probable class:
\begin{equation*}
\hat{y}_i = \underset{k \in \{1,\dots,K\}}{\arg\max}\, \hat{y}_{i,k}.
\end{equation*}
These outputs are the direct result of training the models to minimize task-specific loss functions, as described below. 

\subsubsection*{Objective Functions} 

To generate the predictions used in the metrics above, we optimized models by minimizing standard cross-entropy loss functions, adapting them to the type of classification task and explicitly incorporating class weights to address imbalance. 

\paragraph{Binary Cross-Entropy For Binary Classification Tasks}\mbox{}\\
In binary classification tasks, each sample belongs to one of two classes (e.g., ED disposition: admitted versus discharged). For each sample $i$, the ground-truth label is $y_i \in \{0,1\}$, and the model produces a predicted probability $\hat{y}_i \in [0,1]$ for the positive class through a sigmoid output layer. To correct for imbalance, we applied class-dependent weights $w_{y_i}$ (see Class Imbalance Handling Strategies). The weighted binary cross-entropy (BCE) loss is thus:
\begin{equation*}
\ell_\text{BCE}(y,\hat{y}) \;=\; -\frac{1}{N}\sum_{i=1}^N w_{y_i}\,\big[\,y_i\log(\hat{y}_i) + (1-y_i)\log(1-\hat{y}_i)\,\big].
\end{equation*}

\paragraph{Categorical Cross-Entropy For Multi-Class Classification Tasks}\mbox{}\\
In multi-class classification, each sample belongs to one of $K$ classes (e.g., primary diagnosis at discharge). The ground-truth label for sample $i$ is encoded as a one-hot vector $y_{i,k}$, and the model outputs logits that are passed to a softmax layer to produce class probabilities $\hat{y}_{i,k}$, ensuring the probabilities sum to $1$ in accordance with the law of total probability. As in the binary case, we introduced class-specific weights $w_k$ to mitigate imbalance, with minority classes assigned larger values. The weighted categorical cross-entropy (CCE) loss is therefore:
\begin{equation*}
\ell_\text{CCE}(y,\hat{y}) \;=\; -\frac{1}{N}\sum_{i=1}^N\sum_{k=1}^K w_k \, y_{i,k}\log(\hat{y}_{i,k}).
\end{equation*}

In both cases, as it is a \textit{standard supervised learning setup}, the loss $\ell(y,\hat{y})$ explicitly depends on the ground-truth labels $y$ provided by the dataset, the predicted probabilities $\hat{y}$ produced by the model, and the class weights $\{w_k\}$ derived using a class weighting technique. Incorporating these class weights modifies the effective empirical distribution seen by the optimizer: samples from minority classes are given proportionally greater influence, while those from majority classes are down-weighted. This adjustment reshapes the loss landscape by amplifying gradients associated with underrepresented classes and dampening those from dominant ones, thereby reducing the bias toward majority classes. 

From a statistical learning standpoint, this weighting can be viewed through the lenses of \emph{risk minimization}. The theoretical goal of supervised learning is to minimize the \emph{expected risk}:
\[
R(h) = \mathbb{E}_{(\mathbf{x},y)\sim P}[\,\ell(y,h(\mathbf{x}))\,] = \int \ell(y,h(\mathbf{x})) \, dP(\mathbf{x},y),
\]
where $h: \mathcal{X} \rightarrow \mathcal{Y}$ is a hypothesis function mapping inputs to predicted outputs, and $P$ is the true but unknown data-generating distribution. In practice, training minimizes the \emph{empirical risk}:

\[
\hat{R}(h) = \mathbb{E}_{(\mathbf{x},y)\sim \hat{P}}[\,\ell(y,h(\mathbf{x}))\,]= \frac{1}{N}\sum_{i=1}^N \ell(y_i,h(\mathbf{x}_i)),
\]
which approximates $R(h)$ under the empirical distribution $\hat{P}$ of the observed dataset. In imbalanced settings, however, this empirical distribution does not faithfully represent $P$ or the clinically meaningful importance of classes: majority classes dominate, while minority classes are underrepresented. 

Class weights provide a principled mechanism to re-weight $\hat{R}(h)$ so that it better approximates a desired risk $R_Q(h)$ under some target distribution $Q$. This re-weighting can be \emph{interpreted as analogous to importance sampling}, since the weights $w_k$ adjust the contribution of each class to better reflect $Q$ (e.g., a balanced distribution). In effect, the optimizer no longer minimizes the risk under the raw empirical distribution but under a re-weighted surrogate distribution that emphasizes rare yet clinically critical outcomes. While this promotes more equitable learning across classes, excessively large weights can also inflate gradient variance for minority classes, which may destabilize training underscoring the need for carefully designed weighting strategies.

Intuitively, this process can be viewed as a \emph{transport of distributions}: the observed empirical distribution $\hat{P}$ is skewed toward majority classes, while the desired target distribution $Q$ places greater or proportionate mass on minority or clinically critical classes. Class weights $\{w_k\}$ act as the transport coefficients, redistributing probability mass so that the weighted empirical risk $\hat{R}_w(h)$ becomes a closer surrogate to the theoretical risk $R_Q(h)$. From this perspective, class weighting not only corrects for dataset imbalance but also realigns the optimization objective with the distribution one wishes to learn under, bridging the gap between observed data and theoretical desiderata.

\subsection*{Hyperparameter Optimization}
To ensure optimal and fair comparisons of model performance, systematic hyperparameter optimization was essential. All the models were tuned via Optuna~\cite{akiba2019optuna}, a SOTA Bayesian optimization framework that employs tree-structured Parzen estimator (TPE) sampling to explore hyperparameter spaces efficiently. This approach adaptively focuses computational resources on promising regions on the basis of previous trials, ensuring comprehensive yet efficient optimization across all model architectures.

For each model, we defined comprehensive search spaces covering key hyperparameters that significantly impact performance, as detailed in Appendix~\ref{appendix:hyperparameters}. The optimization process consisted of 100 trials per model-dataset combination, with each trial evaluated via 5-fold cross-validation to ensure robust hyperparameter selection. The objective function was the F1 score on the validation set, which was aligned with our primary evaluation metric. Early stopping was implemented for DL models to prevent overfitting and reduce computational overhead.

Following hyperparameter optimization, the best configuration for each model family was used to train the final models. These optimized models were then evaluated on the held-out test set to generate the results reported in this study. This systematic approach ensures that performance differences between models reflect their inherent capabilities rather than suboptimal hyperparameter choices.
\subsection*{Experimental Setup and Evaluation}
All datasets were split into stratified train, validation, and test partitions (60–20–20\%) to preserve class distributions. Model performance was primarily assessed using the weighted \FOneMacro{}, which is well-suited for imbalanced classification tasks. To evaluate computational efficiency, training times were recorded. Each experiment was repeated for 10 runs with different random seeds. Results are reported as mean~$\pm$~standard deviation, ensuring statistical robustness. All experiments were conducted independently on the MIMIC-IV-ED and eICU datasets.

\section*{Results}
\label{sec:results}
\begin{figure}[!htp]
    \centering
    \includegraphics[width=1.0\linewidth]{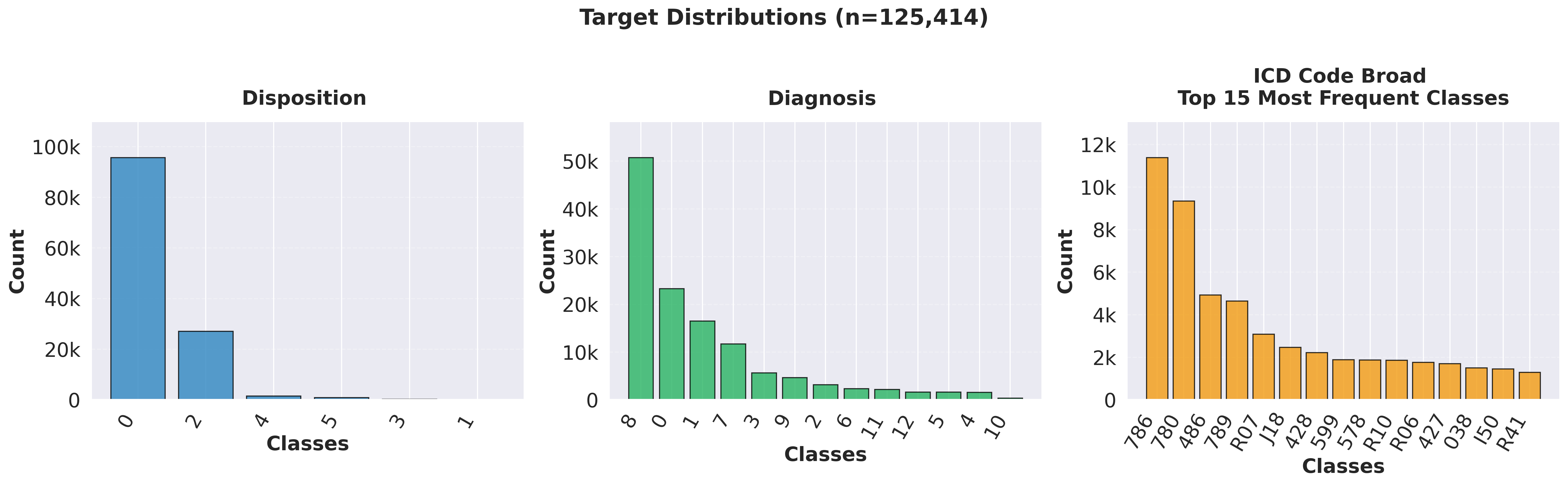}
    \includegraphics[width=1.0\linewidth]{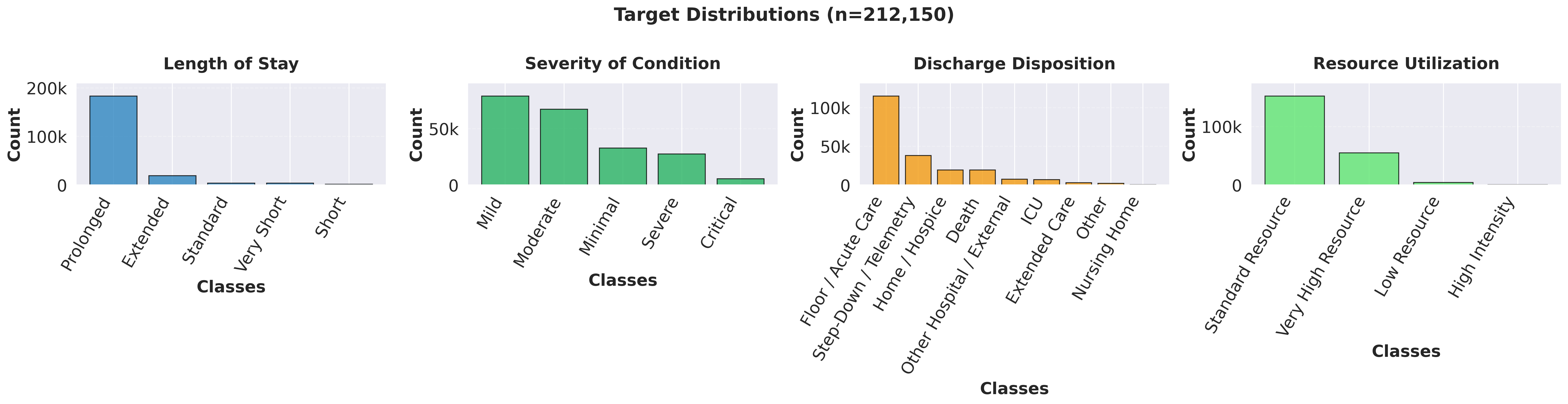}
    \caption{\textbf{Class distribution of  of target outcomes.} Class distribution of target variables in the MIMIC-IV-ED (top row) and eICU (bottom row) datasets. The histograms illustrate the frequency of samples \RW{per class} across various clinical prediction tasks.}
    \label{fig:target_variables_distribution}
\end{figure}
Here, we present the empirical findings. These results are structured to first get a sense of the overall class distributions. Where appropriate, extended analyses and results are provided in the Appendix \ref{appendix:extended_results} and \ref{appendix:eicu_results} to complement our main results.

\subsection*{Class distribution of the two datasets}
\RW{The distributions of the targets across both datasets are presented in Figure~\ref{fig:target_variables_distribution}. The MIMIC-IV-ED dataset exhibits a strong imbalance across all three prediction targets, most especially for diagnosis and ICD groupings, whereas the eICU dataset showed similar skewed patterns for length of stay, severity, discharge disposition, and resource utilization. }



\subsection*{Classifier Performance Comparison}

\begin{figure}[!ht] 
    \centering
    \includegraphics[width=1.0\textwidth]{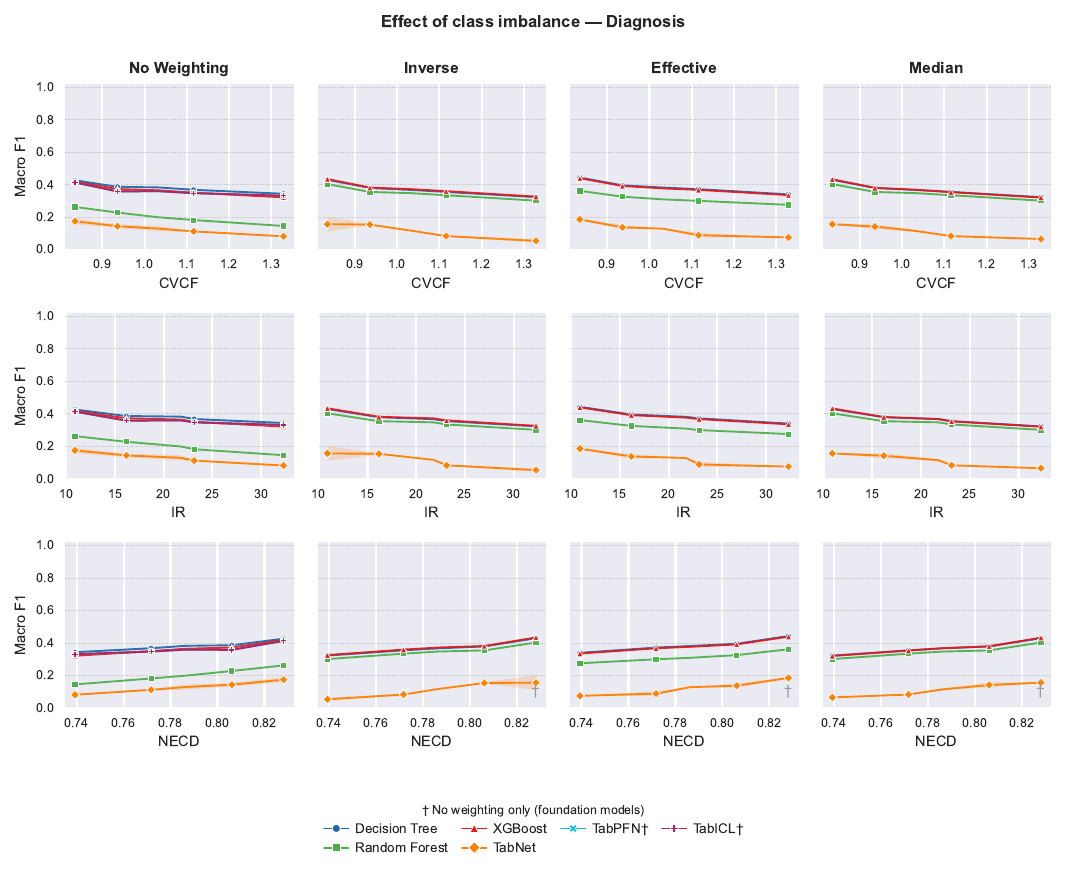} 
    \caption{\textbf{Effect of class imbalance on discharge diagnosis.} 
    Macro F1 performance across varying levels of class imbalance for primary diagnosis prediction. The performance curves for 18 classifier configurations are shown, with the Macro F1 value generally decreasing as imbalance severity increases.}
    \label{fig:class_imbalance_effect_mimic_diagnosis}
\end{figure}

\begin{figure}[!ht] 
    \centering
    \includegraphics[width=1.0\textwidth]{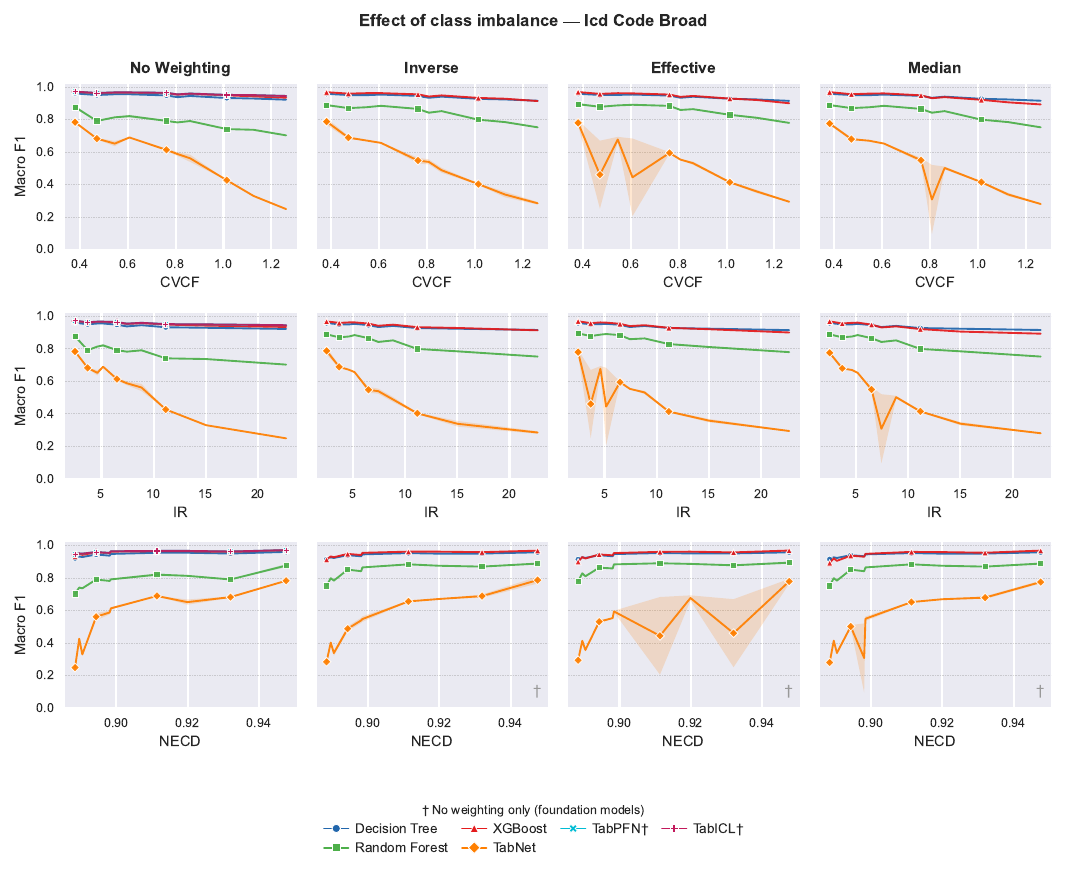} 
    \caption{\textbf{Effect of class imbalance on ICD code prediction.} 
    Macro F1 performance across varying levels of class imbalance for ICD code group prediction. Compared with fine-grained diagnosis prediction, grouped ICD categories reduce label sparsity, and classifiers generally maintain greater stability.}
    \label{fig:class_imbalance_effect_mimic_icd_code_broad}
\end{figure}

\begin{figure}[!ht] 
    \centering
    \includegraphics[width=1.0\textwidth]{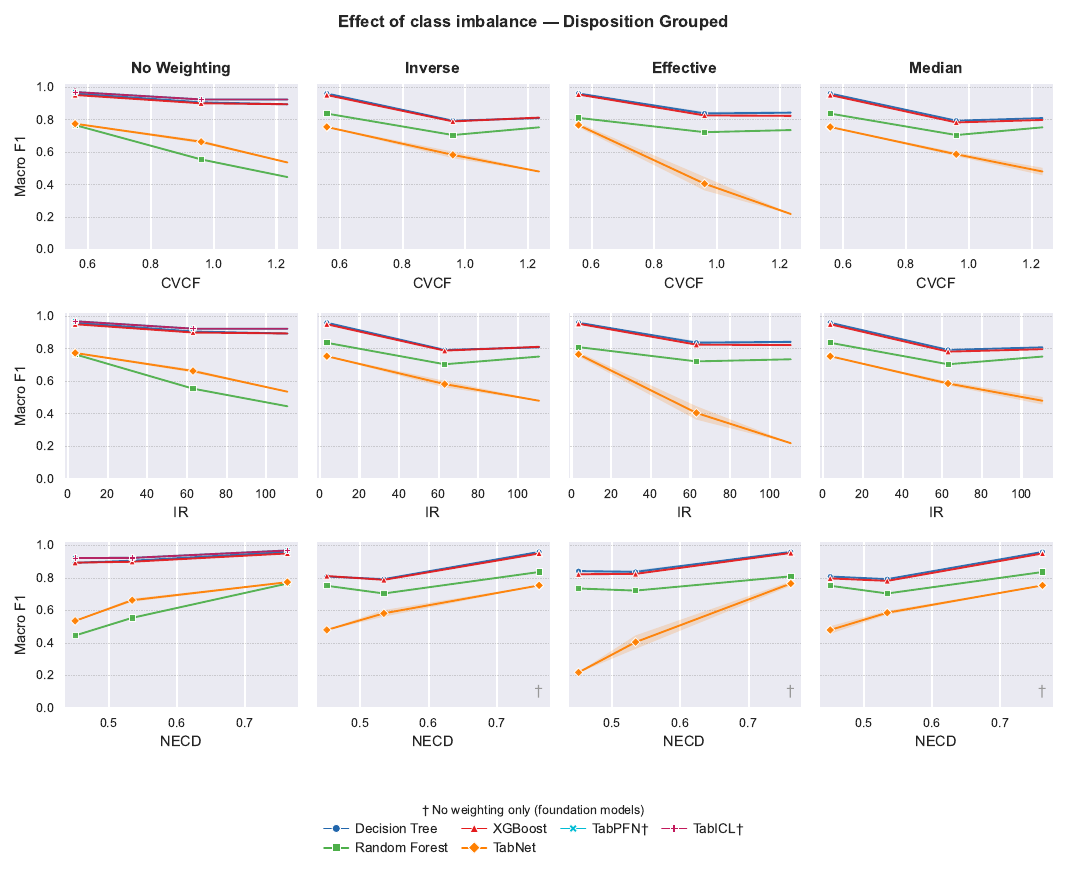} 
    \caption{\textbf{Effect of class imbalance on disposition prediction.} 
    Macro F1 performance across varying levels of class imbalance for patient disposition prediction. The prediction of discharge outcomes shows moderate sensitivity to imbalance, with performance differences depending on the model family and weighting strategy.}
    \label{fig:class_imbalance_effect_mimic_disposition_grouped}
\end{figure}

We evaluated classifier performance using Macro F1 scores across increasing levels of imbalance. Figures~\ref{fig:class_imbalance_effect_mimic_diagnosis}--\ref{fig:class_imbalance_effect_mimic_disposition_grouped} present the results for primary diagnosis, ICD grouping, and discharge disposition prediction in the \MIMIC{} dataset, and Appendix~\ref{eicu_model_performance} reports the corresponding eICU results. We considered 18 classifier configurations spanning 6 model families (DT, RF, XGBoost, TabNet, TabPFN, and TabICL). \RW{For the four trainable models, we evaluated all four imbalance-handling settings resulting in 16 configurations. In addition, the two inference-only foundation models were evaluated exclusively in their pretrained inference regimes without class-weighting strategies, yielding 18 classifier configurations overall.} Macro F1 scores generally declined as imbalance increased, although the magnitude and ordering of the effects varied by task, dataset, and model family.

\begin{figure}[!ht]
    \centering
    \includegraphics[width=1.0\textwidth]{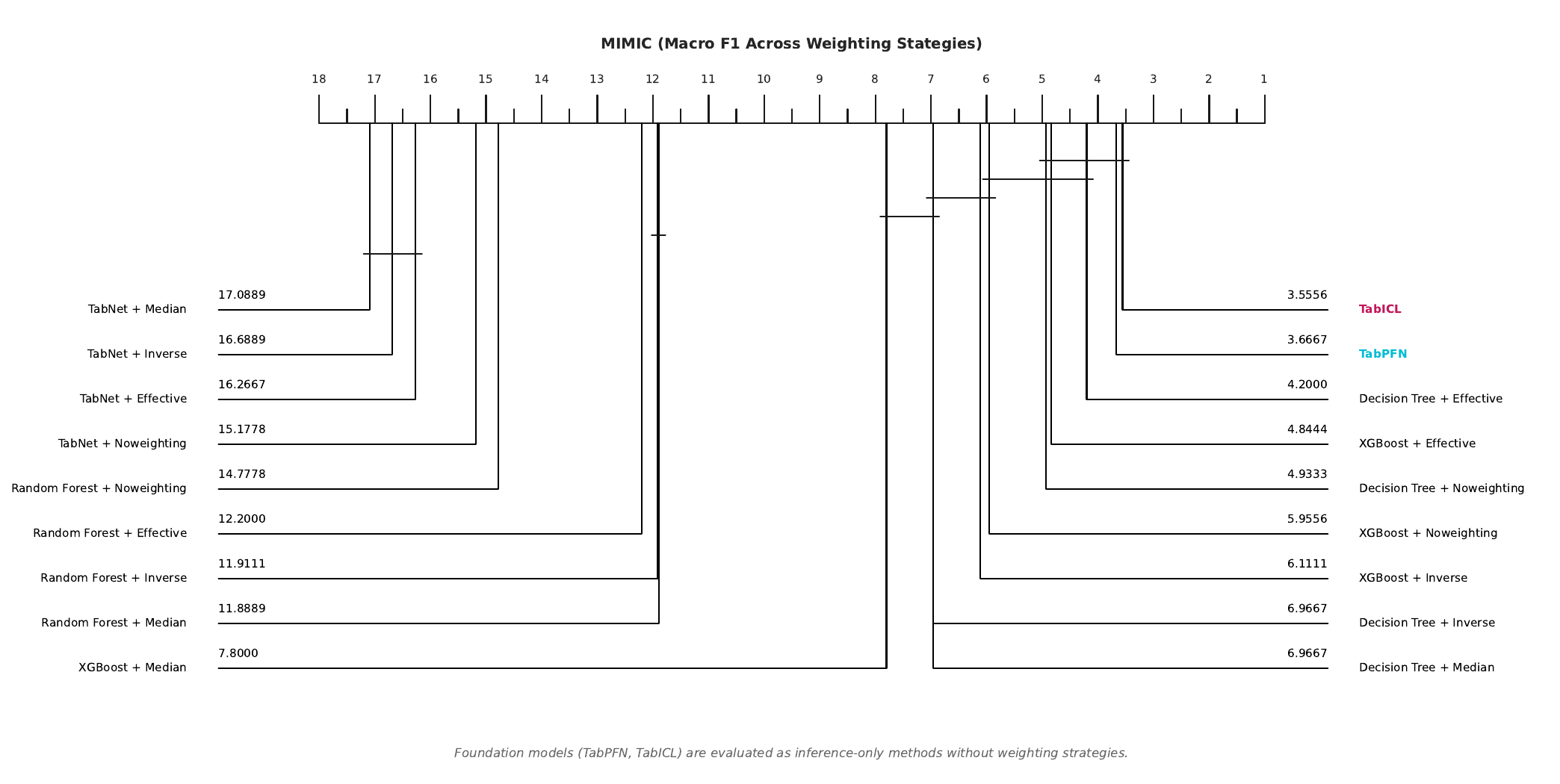}
    \caption{\textbf{Critical difference analysis of classifier performance.} 
    Average ranks of the 18 classifier configurations on the basis of Macro F1 performance across experimental blocks. Lower ranks indicate better predictive performance. The classifiers connected by a horizontal bar are not significantly different from each other according to Wilcoxon signed-rank tests with Holm correction.}
    \label{fig:critical_difference_diagram_mimic_all_classifiers}
\end{figure}

The task-specific plots suggest that primary diagnosis prediction remained the most sensitive to imbalance, reflecting the large number of rare labels in this setting. ICD grouping was comparatively more stable, consistent with the reduced label sparsity created by aggregating diagnoses into broader categories. Disposition prediction showed an intermediate pattern, with imbalance effects remaining visible but less pronounced than in the fine-grained diagnosis task.

Across all tasks, the three imbalance metrics produced broadly consistent degradation patterns. Increases in IR and CVCF, or decreases in NECD (toward 0 from its maximum value of 1 under perfect balance), were associated with lower Macro F1. In the higher-cardinality settings, CVCF was slightly more variable than IR and NECD, which is consistent with its greater sensitivity to distributional spread across moderately rare classes. Overall, the three metrics gave a coherent picture of how predictive performance deteriorates as imbalance becomes more severe.

A Friedman test was used to assess differences in classifier performance across both datasets \RW{(\MIMIC{}: $\chi^2(17, N = 45) = 655.45$, $p = 2.51 \times 10^{-128}$; eICU: $\chi^2(17, N = 32) = 374.83$, $p = 3.34 \times 10^{-69}$). Here, the 17 degrees of freedom correspond to the $k-1$ comparisons among the 18 classifier configurations arising from four trainable model families evaluated under four weighting strategies (16 configurations), together with the two inference-only foundation models (TabPFN and TabICL) evaluated in their pretrained regimes without weighting.} The number of blocks $N$ reflects the distinct prediction tasks crossed with training sample sizes (MIMIC-IV-ED: 3 targets $\times$ 15 filter sizes = 45 blocks; eICU: 4 targets $\times$ 8 filter sizes = 32 blocks), with performance values averaged across 10 experimental runs within each block\RW{.} Post-hoc pairwise comparisons using Wilcoxon signed-rank tests with Holm correction are summarized in Figure~\ref{fig:critical_difference_diagram_mimic_all_classifiers} for \MIMIC{} and Figure~\ref{fig:critical_difference_diagram_eicu_all_classifiers} for eICU (Appendix~\ref{eicu_model_performance}). Overall, the rank-based comparisons did not indicate a single dominant family across both datasets. On \MIMIC{}, the TabICL and TabPFN-based variants attained the strongest average ranks, with XGBoost and DT remaining competitive, whereas TabNet and random forest variants generally ranked lower. On eICU, by contrast, XGBoost variants retained the best overall ranks, followed by TabPFN and TabICL while random forest and decision tree variants occupied intermediate positions and TabNet was less competitive. Taken together, these findings suggest that the relative performance of foundation-style tabular models may depend on the dataset and task setting rather than being universal.

Across both datasets, weighting strategies based on the effective number of samples remained a competitive choice, although their relative advantage varied by model family and task. In several settings they matched or exceeded inverse-frequency and median-frequency weighting, but the expanded comparison does not support a single weighting strategy as uniformly optimal.

Having established these performance differences, we next examine the efficiency of model fitting as dataset size and imbalance scale.

\subsection*{\RW{Model fitting time and compute efficiency characteristics}}

Training efficiency was assessed using both rank-based comparisons and computational time scaling curves. Across the 18 evaluated classifier configurations and experimental blocks, model fitting times differed significantly according to the Friedman test 
\RW{(\MIMIC{}: $\chi^2(17, N = 45) = 48.80$, $p = 6.47 \times 10^{-5}$; eICU: $\chi^2(17, N = 32) = 58.04$, $p = 2.21 \times 10^{-6}$).}

\begin{figure}[!ht]
    \centering
    \includegraphics[width=\textwidth]{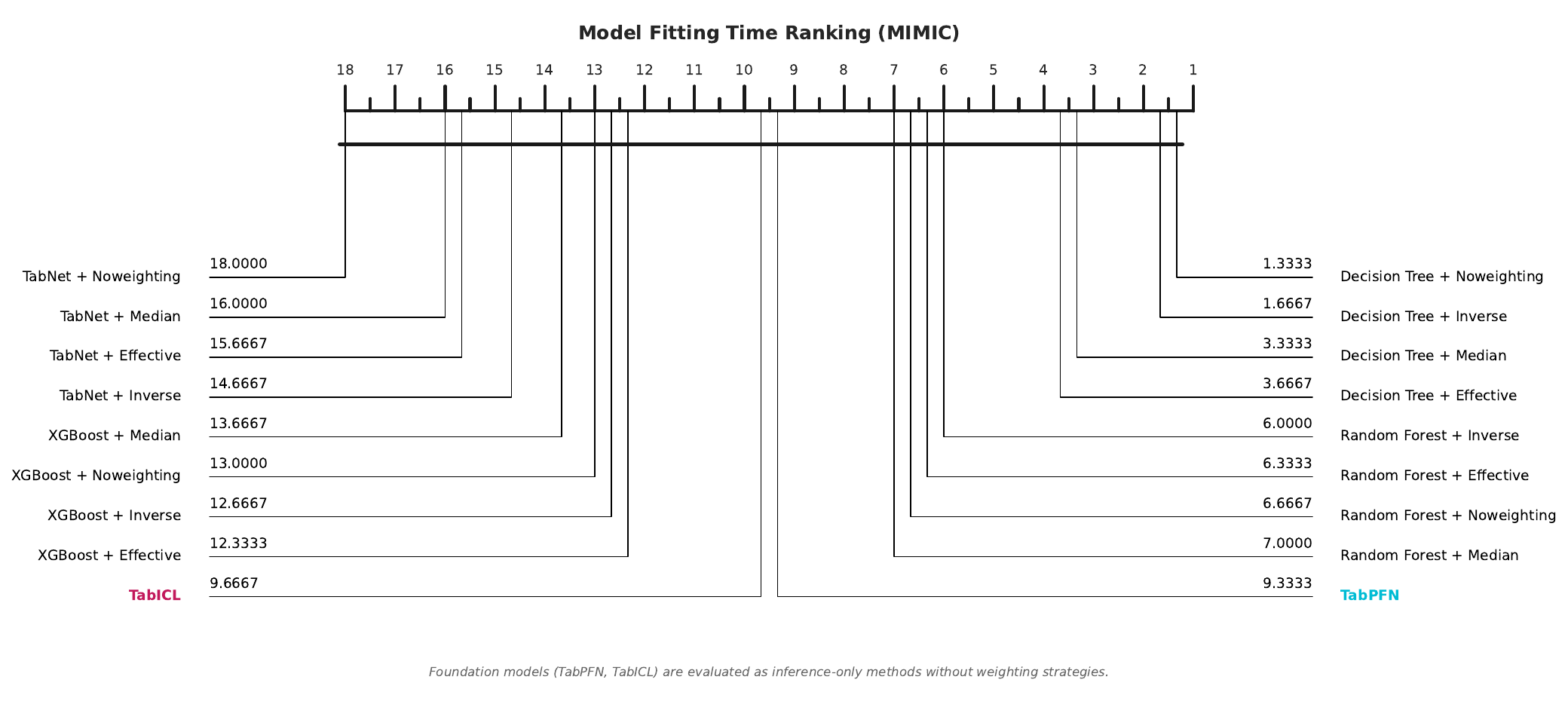}
    \caption{\textbf{Critical difference analysis of classifier training times.}
    Average ranks of 18 classifier configurations across experimental blocks, where each block corresponds to a unique combination of target variable and training set size. Lower ranks indicate faster training. Classifiers connected by a horizontal bar are not significantly different according to Wilcoxon signed-rank tests with Holm correction.}
    \label{fig:critical_difference_diagram_mimic_training_time_with_samples}
\end{figure}

Post-hoc pairwise comparisons using Wilcoxon signed-rank tests with Holm correction (Figure~\ref{fig:critical_difference_diagram_mimic_training_time_with_samples} and Appendix Figure~\ref{fig:critical_difference_diagram_eicu_training_time_with_samples}) indicated a consistent ranking pattern across datasets. Decision Tree variants achieved the lowest average ranks (fastest model fitting times), followed by Random Forest. TabPFN and TabICL occupied intermediate positions in the ranking, reflecting the relatively low time required for inference and support-set conditioning compared to full model optimization procedures. XGBoost variants were second to last in the ranking. And TabNet variants consistently exhibited the highest fitting times.

\RW{It is worth pointing out a key distinguish between two different compute regimes captured in this analysis. For trainable models, model fitting time refers to the wall-clock time required to optimize model parameters on the training data. In contrast, TabPFN and TabICL follow an inference-based paradigm, where no gradient-based training is performed. For these models, the reported fitting time corresponds to the time required to condition on the provided support set. Therefore, these values reflect inference-time adaptation costs rather than conventional training time.}

The scaling curves in Figures~\ref{fig:model_time_performance_mimic_all_targets} (Appendix~\ref{mimic_training_time_plots}) and~\ref{fig:model_time_performance_eicu_all_targets} (Appendix~\ref{eicu:training_time_curves}) further show that differences between model families become more pronounced as the training set size increases. \RW{Decision Tree methods exhibit relatively gradual growth in computational cost with fitting time ranging from 1-3 seconds on the MIMIC-IV-ED database and similars trends on the eICU dataset with training time ranging between 10 and 30 seconds. TabNet show steeper increases, particularly at larger sample sizes. XGBoost exhibited similar trends. Overall, these results suggest that both model architecture and learning paradigm influence computational efficiency, with inference-based foundation models offering a distinct trade-off between amortized training and task-specific adaptation cost. TabPFN and TabICL showed lower compute footprint owning to their forward-pass only setup and prediction on the fly.}

\section*{Discussion}
\label{sec:discussion}
In this section, we interpret our main findings across three principal themes: (i) the associative relationship between imbalance metrics and performance, (ii) the computational scaling behavior of different model families, and (iii) the relative empirical performance of ensemble, conventional deep learning, and tabular foundation models under controlled imbalance conditions. We then consider broader implications for cross-institutional robustness, clinical deployment, and equity. 

\subsection*{Quantifiable Performance Degradation Under Imbalance}
Our experiments revealed consistent, often monotonic relationships between imbalance severity and predictive performance degradation across model families and prediction targets of varying complexity. IR, NECD, and CVCF exhibited strong mutual correlations across all targets and datasets, indicating that these metrics capture fundamentally related aspects of class distribution skew, albeit through varied mathematical formalizations.

IR emphasizes distributional extremes by quantifying the ratio between the most and least frequent classes, NECD captures overall distributional uncertainty on a bounded $[0,1]$ scale (with 1 indicating perfect balance and 0 indicating complete concentration in a single class), and CVCF measures relative dispersion of class frequencies across the full distribution. Despite their distinct formulations, all three metrics were monotonically associated with performance degradation: as imbalance severity increased, reflected by higher IR and CVCF or lower NECD, Macro F1 scores systematically declined. The strong correlations between IR and NECD in particular shows they are measuring fundamentally similar phenomena from complementary perspectives. CVCF showed marginally weaker associations in extremely high-cardinality settings, consistent with its greater sensitivity to moderate frequency variation across many moderately rare classes rather than extremes.

This convergence across metrics strengthens rather than diminishes their practical utility. The fact that three derived measures consistently associated with similar degradation patterns provides robust evidence that class distribution skew is a quantifiable, predictable source of model degradation. The monotonic, predictable relationships we observed suggest that computationally inexpensive imbalance scores could serve as early indicators of expected model robustness prior to full experimental evaluation. While the three metrics are highly correlated, examining multiple metrics provides methodological robustness and allows practitioners to select the most interpretable measure for their context: whether IR for clear min--max intuition, NECD for information-theoretic interpretation on a bounded scale, or CVCF for statistical dispersion. This convergent evidence supports more evidence-based decisions about training strategies and expected performance ranges.

A Friedman test illustrates important differences in classifier performance across both datasets at the level of model families. The extended post-hoc pairwise comparisons across all 18 classifier configurations spanning 6 model families each evaluated under four weighting strategies further clarified the structure of these differences, though the results were not uniform across both datasets.

On MIMIC-IV-ED, the foundation models: TabPFN and TabICL attained the strongest average ranks, with XGBoost remaining competitive, while TabNet and Random Forest variants generally ranked lower. On eICU, by contrast, XGBoost variants consistently achieved the best overall ranks, followed by Random Forest and Decision Tree variants, while TabPFN and TabICL occupied intermediate positions and TabNet was less competitive. These cross-dataset differences are important: they may indicate that no single model family dominates universally, and that the relative ordering of methods including foundation models is sensitive to data and task characteristics. Specifically, the comparatively stronger performance of TabPFN and TabICL on MIMIC-IV-ED relative to eICU may reflect differences in dataset scale, label structure, or distributional properties that interact with the pre-training dynamics of these models.

Traditional tree-based models showed gradual, approximately linear performance decline as imbalance increased, especially for high-cardinality targets. Deep tabular models generally exhibited sharper degradation at more severe imbalance levels, though the magnitude of this difference varied by dataset and task. These cross-dataset degradation patterns suggest that the relationship between model architecture and imbalance robustness could reflects algorithmic properties to a meaningful degree, though this my need further investigation to see how it is generalizes in other settings.

Regarding weighting strategies, models trained without explicit reweighting were often competitive with, and occasionally superior to, those using inverse or median frequency weighting. This likely reflects two factors. First, ensemble methods already mitigate imbalance through recursive partitioning, reducing sensitivity to external reweighting. Second, aggressive weighting of very small classes can amplify gradient noise and destabilize training. In contrast, the effective number of samples scheme~\cite{cui2019class} consistently provided competitive performance across model families and datasets, likely because it moderates minority class influence without overemphasizing extremely rare outcomes. These findings suggest that effective-number weighting represents a robust default, though no single strategy was uniformly optimal across all settings examined.

\subsection*{Empirical Scaling Relationships for Computational Efficiency}
Training time analyses revealed consistent and statistically significant differences in computational efficiency across model families. Post-hoc pairwise comparisons indicated a consistent ordering across datasets: \RW{decision three variants achieved the lowest average training time ranks (fastest), followed by random forest variants}. TabPFN and TabICL occupied intermediate positions in the ranking, reflecting the relatively low time required for inference and support-set conditioning compared to full model optimization procedures. \RW{XGBoost was the penultimate in the ranking.} TabNet variants consistently exhibited the highest training times across configurations. Weighting strategies did not materially alter computational cost for any model family, confirming that efficiency is largely determined by model architecture and learning paradigm rather than class reweighting.

These scaling patterns have practical relevance for deployment in acute care environments, where models may require periodic adaption or retraining as patient populations, clinical practices, or documentation patterns evolve. Our scaling curves show that efficiency differences between model families become more pronounced as dataset size increases: tree-based methods exhibit relatively gradual growth, while neural architectures such as TabNet show steeper increases at larger sample sizes. These empirical scaling relationships provide a basis for capacity planning that goes beyond rough estimates, although deployment-specific factors such as hardware, infrastructure, and update frequency will ultimately determine the practical significance of these differences in any given clinical context.

\RW{
The comparatively favorable fitting times of TabPFN and TabICL reflect their inference-based paradigm: predictions are generated via a pre-trained model without conventional task-specific gradient optimization, substantially reducing per-task computational cost. This represents a fundamentally different operational regime from models requiring full task-specific training, and the efficiency advantage should therefore not be interpreted as reflecting architectural simplicity alone.}

\RW{Specifically, this downstream efficiency is enabled by substantial upstream pre-training investment. TabPFN is trained on approximately 130 million synthetically generated datasets, requiring roughly two weeks of compute on a single node with eight NVIDIA RTX 2080 Ti GPUs. Similarly, TabICL follows a curriculum-driven pre-training pipeline spanning approximately 20 days on three NVIDIA A100 (40GB) GPUs, divided into 16, 3, and 1 days across training stages 1–3, respectively. This highlights an important lifecycle trade-off for clinical deployment: tabular foundation models shift computational burden from decentralized, task-specific training at the hospital or healthcare unit level to centralized large-scale pretraining, enabling lightweight zero-shot or few-shot adaptation at the point of care.
}

\subsection*{Architectural Complexity and Model Selection}

Our extended experimentation across six model families shows that predictive performance in imbalanced clinical tabular learning is not governed by a simple hierarchy of architectural complexity. Instead, it emerges from an interaction between dataset characteristics, task structure, and learning paradigm. Within this landscape, a central finding is that recent tabular foundation models are rapidly narrowing the performance gap with strong classical baselines such as XGBoost.

\RW{
Across all experiments, classical tree-based ensembles, particularly XGBoost, remain highly competitive and often serve as the strongest or most stable baselines. This reinforces the continued relevance of well-engineered classical methods in structured clinical settings, where strong inductive biases and robustness to heterogeneous feature distributions can outweigh gains from increased model complexity.
}

\RW{
In contrast, TabNet consistently underperforms relative to both tree-based ensembles and other neural approaches, while also incurring substantially higher computational cost. This suggests that its sequential attention mechanism provides limited practical benefit for the datasets considered, aligning with prior findings in tabular learning literature~\cite{grinsztajn2022tree, shwartz2021tabular}, particularly in healthcare settings characterized by sparse or partially structured feature interactions.
}

\RW{
A contrasting pattern emerges for tabular foundation models. TabPFN-based and TabICL-based approaches demonstrate increasingly competitive performance across multiple settings. Rather than relying on task-specific gradient-based optimization, these models leverage large-scale pre-training and perform prediction through amortized inference over a support set, enabling rapid adaptation at inference time. On MIMIC-IV-ED, this paradigm achieves performance that is frequently comparable to, and in some cases competitive with, tree-based ensembles. This indicates that pre-training on diverse synthetic task distributions, combined with in-context or support-set conditioning, can increasingly substitute for explicit task-specific optimization in structured clinical domains.
}

\RW{
However, this advantage is not uniformly observed across datasets. On the eICU dataset, no neural or foundation-based model consistently surpasses XGBoost, suggesting that the effectiveness of inference-based tabular foundation models remains sensitive to dataset properties such as institutional heterogeneity, label granularity, and distributional shift across sites. This highlights that, while the gap is narrowing, generalization remains an active area of research.
}

\RW{
Importantly, the compared model families differ fundamentally in how class imbalance is handled. Weighting strategies were applied exclusively to trainable models to modify the empirical risk during optimization. In contrast, TabPFN and TabICL operate in a pretrained inference regime without task-specific retraining or class reweighting. As a result, the comparison reflects two distinct paradigms: explicit risk minimization under reweighted empirical distributions versus amortized inference over pre-trained priors.
}

\RW{
Taken together, these results indicate that tabular foundation models represent an emerging paradigm that is rapidly gaining traction, closing the performance gap with strong classical methods such as XGBoost. This trend suggests that advances in large-scale pre-training and inference-time adaptation are beginning to translate into competitive performance on structured clinical data, even under challenging conditions such as class imbalance.
}

\RW{
Nevertheless, model performance remains strongly context-dependent. Model selection for imbalanced clinical tabular data should therefore be treated as an empirical, dataset-specific problem shaped by trade-offs between predictive performance, computational efficiency, and deployment constraints.
}

\subsection*{Cross-Institutional Generalizability and Its Implications}
The broadly consistent degradation patterns observed across MIMIC-IV-ED (single-centre) and eICU (multi-centre) provide some evidence that the relationship between class imbalance and model performance reflects algorithmic properties rather than purely dataset-specific artefacts. This cross-institutional reproducibility is a meaningful nugget of insight, as healthcare AI systems frequently encounter distribution shift when deployed outside their training environment.

However, the cross-dataset consistency observed here is primarily at the level of degradation \emph{patterns} rather than absolute performance levels or model rankings. The relative ordering of model families differed notably between the two datasets particularly for foundation-based models indicating that cross-institutional consistency cannot be assumed for all model comparisons. Absolute F1 scores varied substantially between datasets and tasks, reflecting differences in patient complexity, outcome prevalence, label granularity, and data quality. Local calibration and validation therefore remain essential even when leveraging evidence from other institutions to inform initial model selection.

These observations have implications for federated learning and multi-institutional AI collaboratives. Evidence that degradation patterns are broadly reproducible may support shared guidelines for model selection under imbalance, but the dataset-dependence of foundation model performance suggests that generalized recommendations should be made cautiously, and that cross-institutional validation ideally in prospective rather than retrospective settings is necessary before strong conclusions can be drawn.

\subsection*{Clinical and Translational Implications for Emergency and Critical Care}
Our results could have potential implications for clinical practice. Ensemble methods specifically XGBoost across both datasets demonstrated comparatively robust performance under class imbalance, which may translate to more reliable identification of rare but clinically important outcomes. In the disposition prediction task, for example, maintaining performance at higher imbalance levels could in principle support more consistent recognition of patients at risk of in-hospital death or requiring urgent transfer. In diagnosis prediction, where many categories are sparsely represented, robustness under imbalance may reduce the likelihood of systematically overlooking uncommon conditions. In the eICU setting, the relatively stable performance of ensembles on disease severity task could in principle support earlier identification of patients who might benefit from intensified monitoring or prioritized resource allocation.

These potential benefits require prospective validation in live clinical environments before any operational conclusions can be drawn. Our analyses are entirely retrospective, and the translation from improved Macro F1 scores on historical data to tangible clinical benefit is not straightforward.

For context, the Macro F1 scores achieved by optimized models on mortality prediction (0.75--0.90) are numerically comparable to published discrimination values for established clinical scoring systems such as APACHE II (AUC 0.80--0.85)~\cite{lee2014validation, badrinath2018comparison} and SOFA (AUC 0.69--0.92)~\cite{minne2008evaluation, bosch2022predictive, asmarawati2022predictive}. For disposition prediction, performance (Macro F1 0.70--0.80) was broadly aligned with reported accuracy for clinician gestalt (60--75\%). These parallels are approximate and should not be interpreted as direct head-to-head comparisons, since the underlying cohorts, covariates, and evaluation metrics differ substantially. They are provided only as contextual benchmarks to situate our results within established clinical practice.

\begin{figure}[!ht]
\centering
\includegraphics[width=0.95\textwidth]{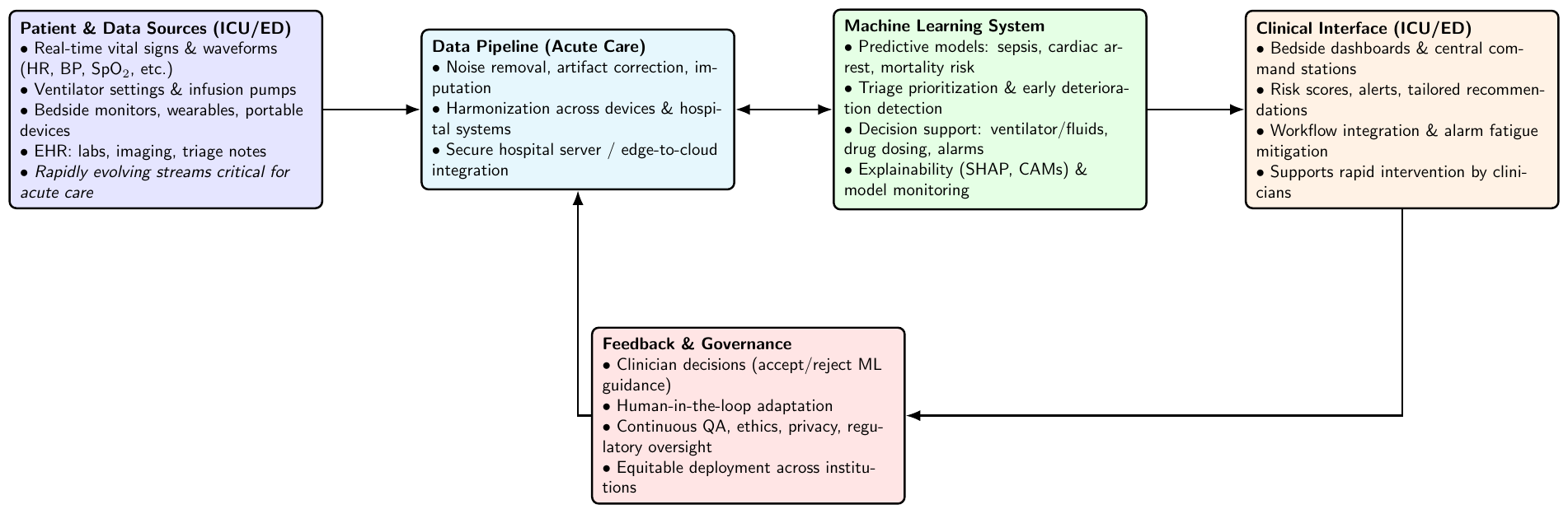}
\caption{\textbf{Conceptual system architecture for AI-enabled clinical decision support.} 
An overview of a ML--enabled clinical decision support system for ICU and ED care. Archival and/ or real-time data streams from monitors, ventilators, infusion pumps, and EHRs feed into predictive models for tasks such as mortality prediction, disposition, and triage prioritization. Model outputs are delivered through clinician-facing dashboards with feedback and governance mechanisms to support safe, equitable, and workflow-aligned deployment.}
\label{fig:icu-ml}
\end{figure}

Predictive accuracy and computational scalability alone are insufficient to guarantee clinical utility. Successful translation requires integration into complex clinical workflows, as illustrated conceptually in Figure~\ref{fig:icu-ml}. Safe and effective deployment will depend on alignment with existing IT infrastructure, and governance structures that ensure continuous monitoring, recalibration, and safety oversight.

Interpretability and clinician trust remain central practical considerations. Tree-based ensembles produce decision rules that can be more readily explained to clinical and regulatory stakeholders than attention-based or foundation-model architectures. Nonetheless, robustness and interpretability alone are insufficient: poorly integrated systems risk contributing to alert fatigue or workflow disruption~\cite{ancker2017effects, chaparro2022clinical}. Human-centred design, iterative clinician feedback, and organizational change management must accompany any algorithmic development effort.

Finally, equity considerations are inseparable from technical ones. Class imbalance often reflects underlying disparities in disease prevalence or access to care, and models that degrade sharply under imbalance risk amplifying such inequities. The relatively stable performance of XGBoost and tabular foundation models across imbalance levels may help mitigate this risk to a degree, though overall robustness to class imbalance does not guarantee equitable performance within demographic subgroups. Their reproducibility across datasets further supports the potential for federated or collaborative approaches, although the equity implications of cross-institutional deployment remain to be systematically assessed.

To sum up, these observations highlight that model choice in clinical AI should consider robustness to imbalance, computational scalability, interpretability, and governance requirements as jointly relevant criteria, not accuracy alone.

\subsection*{Operational Deployment and Implementation Considerations}

Successful clinical deployment requires addressing operational, methodological, and regulatory considerations that extend beyond aggregate predictive performance. In this study, we observed that model behavior is strongly shaped not only by predictive architecture, but also by the underlying learning paradigm. Classical trainable models and inference-based foundation models introduce fundamentally different deployment trade-offs that clinical stakeholders must consider carefully.

Tree-based ensemble methods such as XGBoost and Random Forest demonstrated consistently strong predictive performance while maintaining comparatively favorable computational scaling characteristics. Their efficiency facilitates more frequent retraining cycles as patient populations, documentation practices, and clinical workflows evolve over time. This is operationally relevant in acute care environments where \textit{concept drift} can emerge rapidly and periodic recalibration may be necessary to preserve reliability.

\RW{
Across weighting strategies, no single approach was uniformly optimal across all datasets, prediction tasks, and model families. Nevertheless, several consistent empirical tendencies emerged that may help guide practical deployment decisions. Weighting based on the effective number of samples generally remained a comparatively robust default across a wide range of imbalance conditions, particularly in settings with moderate-to-severe imbalance and higher class cardinality. By contrast, tree-based ensemble methods frequently remained competitive even without explicit reweighting, suggesting that their recursive partitioning structure already provides a degree of robustness to imbalance. More aggressive weighting schemes, such as inverse-frequency weighting, occasionally introduced instability for extremely rare classes, likely due to amplification of gradient noise and sensitivity to small sample fluctuations.
}

\RW{
These findings suggest that weighting strategy selection should be treated as a dataset- and task-dependent operational decision rather than a universally fixed preprocessing choice. In practice, effective-number weighting may represent a reasonable starting point when imbalance severity is substantial or minority-class preservation is clinically important, whereas simpler unweighted training may remain sufficient for some ensemble-based methods under moderate imbalance conditions. Importantly, these observations should be interpreted as empirical heuristics derived from the present experiments rather than prescriptive optimization rules applicable to all clinical tabular learning settings.
}

\RW{
By contrast, TabPFN v2.6 and TabICL operate under an inference-based paradigm in which downstream tasks are solved through amortized inference over a support set rather than conventional task-specific optimization. Consequently, the reported computational cost in this study reflects downstream adaptation and inference time only, not the substantial upstream pre-training cost required to construct these models. This distinction is important for practical interpretation: while foundation models offer low per-task deployment cost and rapid adaptation, this efficiency is enabled by a large centralized pre-training investment that may remain inaccessible to most healthcare institutions.
}

\RW{
The practical implications of this trade-off depend on the intended deployment scenario. In environments requiring frequent local retraining, institutional customization, or transparent optimization pipelines, classical ensemble methods may remain preferable because they can be retrained directly on local data with modest computational resources. Conversely, in settings where rapid deployment across many related tasks is prioritized, pretrained foundation models may provide operational advantages through zero-shot or few-shot adaptation without repeated optimization.
}

Importantly, weighting strategies evaluated in this study apply only to trainable models, where class weights explicitly modify the optimization objective during empirical risk minimization. \RW{TabPFN and TabICL were evaluated strictly in their pretrained inference regimes without task-specific reweighting or retraining.} As such, comparisons between trainable and inference-based approaches should be interpreted as comparisons between distinct learning paradigms rather than interchangeable optimization procedures.

Integration with existing hospital IT infrastructure presents additional technical and workflow challenges. Models must interface with heterogeneous EHR systems, comply with institutional governance policies, and provide outputs compatible with clinical decision support workflows. Tree-based methods retain practical advantages in this context due to their relative interpretability, lower infrastructure requirements, and more established regulatory familiarity compared with large pretrained neural systems.

Alert fatigue remains a major deployment barrier in clinical AI systems~\cite{ancker2017effects, chaparro2022clinical}. Although our cross-institutional experiments suggest some consistency in performance patterns across datasets, prospective validation remains necessary before assuming transferability of operating thresholds between institutions. Governance frameworks should therefore include continuous performance monitoring, recalibration procedures, and clear human override mechanisms.

\RW{
Taken together, these findings suggest that model selection in imbalanced clinical tabular learning should not be framed solely as a question of predictive accuracy. Instead, practical deployment requires balancing predictive robustness, computational efficiency, retraining flexibility, interpretability, infrastructure constraints, and governance requirements. Within this broader operational landscape, tabular foundation models represent a promising but still evolving paradigm whose advantages appear increasingly competitive, yet remain context-dependent across clinical environments.
}

\textbf{\RW{Decision Guidance for Practitioners}}

\RW{Given these findings, we have developed a heuristics-based structured decision framework for model and class weighting strategy selection in imbalanced clinical tabular prediction tasks, summarized in Table~\ref{tab:framework}.}

\RW{Several empirical patterns can be translated into practical decision rules. On single-center datasets with moderate label cardinality (e.g., MIMIC-IV-ED), tabular foundation models~(TabPFNv2.6, TabICL) are strong default choices, particularly when minimal task-specific training is desired and inference-time efficiency is a priority. In contrast, under multi-center or institutionally heterogeneous conditions (e.g., eICU), gradient-boosted tree models (XGBoost) should be preferred, as they consistently provided the most reliable performance across imbalance levels.}

\RW{TabNet is not recommended for highly imbalanced clinical tabular settings due to consistently inferior predictive performance and higher computational cost relative to both tree-based and foundation model alternatives.}

\RW{For imbalance handling, effective number of samples weighting~\cite{cui2019class} is a strong default choice across most settings and model families. Tree-based ensembles often remain competitive without explicit reweighting, indicating inherent robustness to moderate imbalance. In contrast, inverse-frequency weighting should be used cautiously in high-cardinality or extremely rare-class settings due to observed instability effects. Median-frequency weighting shows more variable behavior and does not consistently outperform the other strategies across datasets or model families.}

\RW{From a computational vantage point, decision trees and random forests provide the most efficient training and retraining characteristics, making them suitable for frequent local updates and resource-constrained settings. TabPFN and TabICL offer minimal per-task adaptation cost under their downstream inference-based paradigm, although this efficiency reflects substantial upstream pretraining data and compute requirements that should be considered when evaluating deployment feasibility. XGBoost occupies an intermediate position, combining strong predictive performance with practical retraining flexibility on local clinical data.}

\RW{These rules are intended as structured empirical guidance. Model selection in clinical deployment should still incorporate external validation, prospective evaluation, and monitoring under distribution shift.}

\begin{table}[!ht]
\centering
\caption{Empirically derived model and weighting strategy selection guidance for imbalanced clinical tabular prediction tasks.}
\label{tab:framework}
\small
\begin{tabular}{p{3.6cm} p{3.0cm} p{3.0cm} p{4.0cm}}
\toprule
\textbf{Scenario} & \textbf{Recommended model family} & \textbf{Recommended weighting strategy} & \textbf{Notes} \\
\midrule

High imbalance + high label cardinality & XGBoost & Effective number of samples, or none & Strongest and most stable ranks across both datasets; gradual degradation under increasing imbalance \\[4pt]

Moderate imbalance + single-center structured EHR data & Tabular foundation models (TabPFN v2.6, TabICL) & N/A (inference-only) & Competitive or leading performance on MIMIC-IV-ED; no task-specific retraining required \\[4pt]

Multi-center or institutionally heterogeneous data (any imbalance level) & XGBoost & Effective number of samples, or none & Foundation models less stable on eICU; XGBoost consistently most reliable across settings \\[4pt]

Frequent local retraining required or strong interpretability/regulatory constraints & Decision Tree or Random Forest & Effective number of samples, or none & Fastest training times; Random Forest more stable than single trees under imbalance \\[4pt]

Rapid deployment across many related tasks with centralized infrastructure & Tabular foundation models (TabPFN v2.6, TabICL) & N/A (inference-only) & Lowest per-task adaptation cost; requires substantial upstream pretraining investment \\[4pt]

Any high-imbalance clinical tabular setting where efficiency dominates accuracy & TabNet is not recommended & --- & Consistently weakest performance-to-cost trade-off across both datasets \\
\bottomrule
\end{tabular}
\end{table}

\subsection*{Equity, Fairness, and Bias Considerations}
Class imbalance in healthcare data is not merely a statistical property; it often reflects underlying disparities in disease prevalence, healthcare access, diagnostic practices, or clinical recognition. As a result, class imbalance and algorithmic fairness are related challenges. A model that performs robustly under skewed class distributions may still exhibit clinically significant disparities across demographic or socioeconomic subgroups.

\RW{
This distinction is particularly important because aggregate metrics such as accuracy can obscure poor performance on clinically critical minority outcomes. To combat this limitation, this study primarily reports Macro F1-score, which weights all classes equally irrespective of prevalence and therefore provides greater sensitivity to minority-class degradation than weighted or micro-averaged metrics. Nevertheless, robust macro-level performance does not guarantee equitable subgroup behavior.
}

Minority outcome classes in our prediction tasks, such as adverse disposition outcomes or rare diagnoses, may themselves be distributed unevenly across patient populations defined by race, sex, socioeconomic status, geography, or insurance status. Consequently, models that appear robust under aggregate imbalance conditions may still propagate or amplify disparities when deployed in practice.

\RW{
Recent work has further emphasized that fairness-aware learning should account not only for prediction disparities but also for uncertainty and representation bias within the optimization process itself~\cite{incremona2025differentiable}. From this perspective, class imbalance mitigation and fairness mitigation should not be treated as interchangeable objectives. Reweighting strategies designed to improve minority-class recognition may improve predictive robustness without necessarily improving demographic equity, and in some cases may even shift error distributions unevenly across subgroups.
}

A practical roadmap for future fairness evaluation therefore include: (i) reporting subgroup-specific discrimination metrics such as AUC-ROC, AUPRC, precision, and recall stratified by demographic variables; (ii) evaluating calibration and threshold-dependent fairness criteria such as equal opportunity and false-negative parity; (iii) analyzing intersectional strata (e.g., sex $\times$ age or race/ethnicity $\times$ insurance status) to identify compounded disparities; and (iv) assessing distributional drift and subgroup performance continuously after deployment.

\RW{
The differing computational characteristics of the evaluated model families can also influence fairness in practice. Efficient classical models can support more frequent local retraining and recalibration across demographic strata, whereas large pretrained foundation models may depend on centralized updates that are less adaptable to institution-specific population shifts. This introduces an additional governance consideration: deployment feasibility and update flexibility can themselves shape downstream equity outcomes.
}

Technical fairness evaluations must ultimately be interpreted within the broader context of health equity, since algorithmic metrics alone cannot fully capture structural and institutional drivers of disparity~\cite{donini2018empirical, oneto2020general, abramoff2023considerations, sikstrom2022conceptualising}. Systemic risks remain if deployment decisions are driven primarily by institutional resources rather than clinical appropriateness. Resource-constrained settings may systematically receive less adaptable or less well-supported systems, potentially reinforcing existing inequities in access to clinical AI technologies.

\subsection*{Contextualization With Existing Literature}
This study builds on and extends prior research underscoring the effectiveness of tree-based methods on tabular data~\cite{shwartz2021tabular, grinsztajn2022tree}. Our contribution extends this work by systematically evaluating a broader set of model families, including two recent tabular foundation models, under controlled imbalance conditions and by quantifying computational scaling relationships specifically in ED and ICU clinical contexts.

Whereas earlier studies of deep learning and traditional methods on healthcare data have often been restricted to specific clinical domains or single datasets~\cite{rajkomar2018scalable, Choi2017}, our cross-institutional validation provides a broader evidence base. In particular, the inclusion of TabPFN v2.6~\cite{grinsztajn2025tabpfn} and TabICL~\cite{qu2025tabicl} reflects the current state of the tabular learning landscape. TabPFN v2.6 represents a substantially more capable and efficient version of the original TabPFN framework, incorporating a distilled variant suitable for large-scale deployment, making its inclusion particularly relevant for clinical feasibility assessments. The dataset-dependent behavior of these foundation models, competitive on MIMIC-IV-ED but more modest on eICU, adds nuance to existing discussions about when and where pre-trained tabular models provide practical advantages.

\RW{
In contrast to earlier neural approaches for tabular data, recent foundation models such as TabPFN v2.6 and TabICL increasingly reduce the gap to strong classical baselines. This shift is particularly evident in structured clinical settings, where carefully engineered tree-based ensembles like XGBoost remain highly competitive but are no longer universally dominant. This trend reflects rapid progress in large-scale pre-training and inference-time adaptation strategies for tabular learning.
}

\subsection*{Limitations and Future Research Directions}

While the utilization of TabPFN v2.6~\cite{grinsztajn2025tabpfn} and TabICL~\cite{qu2025tabicl} substantially broadens the scope of the current study, the tabular learning landscape continues to evolve rapidly. Architectures such as RealTabPFN~\cite{garg2025real}, TabDPT~\cite{ma2024tabdpt}, and other recently emerging models highlighted in benchmarks such as TabArena were not evaluated here. Likewise, alternative gradient boosting frameworks including LightGBM~\cite{ke2017lightgbm} and CatBoost~\cite{prokhorenkova2018catboost} were excluded in favor of a single representative boosting implementation to keep the analysis focused on broader model families rather than exhaustive within-family comparisons. Although this choice improves interpretability of the comparisons, it limits the extent to which conclusions can be generalized to every individual implementation or newly proposed architecture.

The scope of this analysis was on structured EHR data. Clinical AI is increasingly moving toward multimodal learning paradigms that combine structured records with medical imaging, clinical notes, genomics, and physiological time series. Under these richer modalities, deep learning and foundation-based approaches may exhibit different scaling behavior and potentially stronger relative advantages than observed in the present study.

Class imbalance mitigation was decidedly limited to weighting strategies that can be applied consistently across trainable model families, allowing comparisons under a unified experimental framework. This choice may nevertheless disadvantage certain neural architectures that could benefit from more specialized imbalance-aware techniques such as focal loss~\cite{lin2017focal}, curriculum sampling, progressive resampling, or self-supervised pre-training. In addition, the behavior of inference-based foundation models under alternative imbalance mitigation or calibration strategies remains insufficiently understood and warrants further investigation.

Furthermore, the experiments were conducted on static retrospective datasets and therefore do not capture the temporal dynamics of real clinical environments. In practice, patient populations, disease prevalence, documentation practices, and clinical workflows evolve continuously over time. As a result, the performance patterns and computational scaling relationships reported here may differ under streaming or continuously updated deployment conditions. Predictive improvements observed on historical datasets do not necessarily translate into clinical utility once issues such as workflow integration, clinician trust, alert fatigue, institutional governance, and regulatory oversight are considered. Prospective studies embedded within real clinical workflows remain necessary before operational conclusions can be drawn.

\RW{
The present analysis also focused primarily on overall predictive robustness rather than subgroup-specific fairness. Although Macro F1-score provides greater sensitivity to minority-class performance than aggregate accuracy metrics, it does not directly assess demographic equity. Whether the robustness advantages observed for certain model families translate into equitable performance across patient populations defined by race, sex, age, geography, or socioeconomic status remains an important open question. Addressing this will require dedicated subgroup-level analyses, calibration studies, and fairness-aware evaluation protocols.
}

Both MIMIC-IV-ED and eICU originate from academic medical centers in the United States. The extent to which these findings generalize to community hospitals, non-U.S. healthcare systems, or resource-constrained clinical environments remains uncertain and should be validated more broadly.

Future work could extend these findings in several directions, including prospective validation in live clinical settings, broader fairness and calibration analyses, benchmarking against newly emerging tabular foundation models, and expansion toward multimodal clinical AI systems. Another promising direction lies in hybrid approaches that combine the robustness and efficiency of classical ensemble methods with the representational flexibility and adaptation capabilities of pretrained foundation models.

\section*{Conclusion}
\RW{Robustness to class imbalance and computational efficiency are fundamental requirements for clinical AI systems deployed in emergency and intensive care environments, where rare outcomes are often the most clinically consequential and operational constraints directly influence feasibility of deployment. This study provides a systematic empirical evaluation of these challenges across two large-scale clinical datasets, multiple prediction tasks, six model families, and complementary imbalance quantification metrics.}

\RW{Several key findings emerge from this analysis. First, strong classical ensemble methods, particularly XGBoost, remain highly competitive across a wide range of imbalanced clinical tabular prediction tasks, reinforcing their continued practical relevance for healthcare AI. Second, recent tabular foundation models including TabPFN v2.6 and TabICL demonstrate that inference-based paradigms are rapidly narrowing the performance gap with established classical approaches, particularly on certain structured clinical datasets. Although these models do not consistently outperform ensemble methods across all settings, the results suggest that advances in large-scale pre-training and inference-time adaptation are beginning to translate into practically competitive performance under realistic clinical imbalance conditions.}

\RW{The study further shows that predictive performance alone is insufficient for evaluating clinical AI systems. Differences in computational scaling, retraining flexibility, interpretability, deployment constraints, and fairness considerations all materially influence the suitability of a model family for real-world healthcare settings. Importantly, trainable models and inference-based foundation models represent fundamentally different optimization paradigms, and their reported computational efficiencies should therefore be interpreted within the context of distinct lifecycle trade-offs between centralized pre-training and downstream task adaptation.}

\RW{Taken together, the findings indicate that model selection for imbalanced clinical tabular data is inherently context-dependent rather than governed by a universal architectural hierarchy. More broadly, the results position tabular foundation models as an emerging and increasingly competitive paradigm for structured clinical learning while simultaneously reaffirming the enduring strength and practicality of carefully engineered classical ensemble methods under realistic clinical constraints.}
\bibliography{sample}

\begin{thebibliography}{10}
\urlstyle{rm}
\expandafter\ifx\csname url\endcsname\relax
  \def\url#1{\texttt{#1}}\fi
\expandafter\ifx\csname urlprefix\endcsname\relax\def\urlprefix{URL }\fi
\expandafter\ifx\csname doiprefix\endcsname\relax\def\doiprefix{DOI: }\fi
\providecommand{\bibinfo}[2]{#2}
\providecommand{\eprint}[2][]{\url{#2}}

\bibitem{huang2010impact}
\bibinfo{author}{Huang, Q.}, \bibinfo{author}{Thind, A.}, \bibinfo{author}{Dreyer, J.~F.} \& \bibinfo{author}{Zaric, G.~S.}
\newblock \bibinfo{journal}{\bibinfo{title}{The impact of delays to admission from the emergency department on inpatient outcomes}}.
\newblock {\emph{\JournalTitle{BMC emergency medicine}}} \textbf{\bibinfo{volume}{10}}, \bibinfo{pages}{16} (\bibinfo{year}{2010}).

\bibitem{carayon2014human}
\bibinfo{author}{Carayon, P.}, \bibinfo{author}{Xie, A.} \& \bibinfo{author}{Kianfar, S.}
\newblock \bibinfo{journal}{\bibinfo{title}{Human factors and ergonomics as a patient safety practice}}.
\newblock {\emph{\JournalTitle{BMJ Quality \& Safety}}} \textbf{\bibinfo{volume}{23}}, \bibinfo{pages}{196--205} (\bibinfo{year}{2014}).

\bibitem{weigl2012association}
\bibinfo{author}{Weigl, M.}, \bibinfo{author}{M{\"u}ller, A.}, \bibinfo{author}{Vincent, C.}, \bibinfo{author}{Angerer, P.} \& \bibinfo{author}{Sevdalis, N.}
\newblock \bibinfo{journal}{\bibinfo{title}{The association of workflow interruptions and hospital doctors' workload: a prospective observational study}}.
\newblock {\emph{\JournalTitle{BMJ quality \& safety}}} \textbf{\bibinfo{volume}{21}}, \bibinfo{pages}{399--407} (\bibinfo{year}{2012}).

\bibitem{johnson2016machine}
\bibinfo{author}{Johnson, A.~E.} \emph{et~al.}
\newblock \bibinfo{journal}{\bibinfo{title}{Machine learning and decision support in critical care}}.
\newblock {\emph{\JournalTitle{Proceedings of the IEEE}}} \textbf{\bibinfo{volume}{104}}, \bibinfo{pages}{444--466} (\bibinfo{year}{2016}).

\bibitem{komorowski2018artificial}
\bibinfo{author}{Komorowski, M.}, \bibinfo{author}{Celi, L.~A.}, \bibinfo{author}{Badawi, O.}, \bibinfo{author}{Gordon, A.~C.} \& \bibinfo{author}{Faisal, A.~A.}
\newblock \bibinfo{journal}{\bibinfo{title}{The artificial intelligence clinician learns optimal treatment strategies for sepsis in intensive care}}.
\newblock {\emph{\JournalTitle{Nature medicine}}} \textbf{\bibinfo{volume}{24}}, \bibinfo{pages}{1716--1720} (\bibinfo{year}{2018}).

\bibitem{esteva2019guide}
\bibinfo{author}{Esteva, A.} \emph{et~al.}
\newblock \bibinfo{journal}{\bibinfo{title}{A guide to deep learning in healthcare}}.
\newblock {\emph{\JournalTitle{Nature medicine}}} \textbf{\bibinfo{volume}{25}}, \bibinfo{pages}{24--29} (\bibinfo{year}{2019}).

\bibitem{topol2019high}
\bibinfo{author}{Topol, E.~J.}
\newblock \bibinfo{journal}{\bibinfo{title}{High-performance medicine: the convergence of human and artificial intelligence}}.
\newblock {\emph{\JournalTitle{Nature medicine}}} \textbf{\bibinfo{volume}{25}}, \bibinfo{pages}{44--56} (\bibinfo{year}{2019}).

\bibitem{rajpurkar2022ai}
\bibinfo{author}{Rajpurkar, P.}, \bibinfo{author}{Chen, E.}, \bibinfo{author}{Banerjee, O.} \& \bibinfo{author}{Topol, E.~J.}
\newblock \bibinfo{journal}{\bibinfo{title}{Ai in health and medicine}}.
\newblock {\emph{\JournalTitle{Nature medicine}}} \textbf{\bibinfo{volume}{28}}, \bibinfo{pages}{31--38} (\bibinfo{year}{2022}).

\bibitem{miotto2018deep}
\bibinfo{author}{Miotto, R.}, \bibinfo{author}{Wang, F.}, \bibinfo{author}{Wang, S.}, \bibinfo{author}{Jiang, X.} \& \bibinfo{author}{Dudley, J.~T.}
\newblock \bibinfo{journal}{\bibinfo{title}{Deep learning for healthcare: review, opportunities and challenges}}.
\newblock {\emph{\JournalTitle{Briefings in bioinformatics}}} \textbf{\bibinfo{volume}{19}}, \bibinfo{pages}{1236--1246} (\bibinfo{year}{2018}).

\bibitem{jiang2017artificial}
\bibinfo{author}{Jiang, F.} \emph{et~al.}
\newblock \bibinfo{journal}{\bibinfo{title}{Artificial intelligence in healthcare: past, present and future}}.
\newblock {\emph{\JournalTitle{Stroke and vascular neurology}}} \textbf{\bibinfo{volume}{2}} (\bibinfo{year}{2017}).

\bibitem{He2009}
\bibinfo{author}{He, H.} \& \bibinfo{author}{Garcia, E.~A.}
\newblock \bibinfo{journal}{\bibinfo{title}{Learning from imbalanced data}}.
\newblock {\emph{\JournalTitle{IEEE Transactions on Knowledge and Data Engineering}}} \textbf{\bibinfo{volume}{21}}, \bibinfo{pages}{1263--1284}, \doiprefix\url{10.1109/TKDE.2008.239} (\bibinfo{year}{2009}).

\bibitem{Branco2016}
\bibinfo{author}{Branco, P.}, \bibinfo{author}{Torgo, L.} \& \bibinfo{author}{Ribeiro, R.~P.}
\newblock \bibinfo{journal}{\bibinfo{title}{A survey of predictive modeling on imbalanced domains}}.
\newblock {\emph{\JournalTitle{ACM Computing Surveys}}} \textbf{\bibinfo{volume}{49}}, \bibinfo{pages}{31:1--31:50}, \doiprefix\url{10.1145/2907071} (\bibinfo{year}{2016}).

\bibitem{lecun2015deep}
\bibinfo{author}{LeCun, Y.}, \bibinfo{author}{Bengio, Y.} \& \bibinfo{author}{Hinton, G.}
\newblock \bibinfo{journal}{\bibinfo{title}{Deep learning}}.
\newblock {\emph{\JournalTitle{Nature}}} \textbf{\bibinfo{volume}{521}}, \bibinfo{pages}{436--444} (\bibinfo{year}{2015}).

\bibitem{krizhevsky2012imagenet}
\bibinfo{author}{Krizhevsky, A.}, \bibinfo{author}{Sutskever, I.} \& \bibinfo{author}{Hinton, G.~E.}
\newblock \bibinfo{title}{Imagenet classification with deep convolutional neural networks}.
\newblock In \emph{\bibinfo{booktitle}{Advances in Neural Information Processing Systems}}, vol.~\bibinfo{volume}{25} (\bibinfo{year}{2012}).

\bibitem{silver2017mastering}
\bibinfo{author}{Silver, D.} \emph{et~al.}
\newblock \bibinfo{journal}{\bibinfo{title}{Mastering the game of go without human knowledge}}.
\newblock {\emph{\JournalTitle{Nature}}} \textbf{\bibinfo{volume}{550}}, \bibinfo{pages}{354--359} (\bibinfo{year}{2017}).

\bibitem{Arik2021}
\bibinfo{author}{Arik, S.~O.} \& \bibinfo{author}{Pfister, T.}
\newblock \bibinfo{journal}{\bibinfo{title}{Tabnet: Attentive interpretable tabular learning}}.
\newblock {\emph{\JournalTitle{Proceedings of the AAAI Conference on Artificial Intelligence}}} \textbf{\bibinfo{volume}{35}}, \bibinfo{pages}{6679--6687}, \doiprefix\url{10.1609/aaai.v35i8.16665} (\bibinfo{year}{2021}).

\bibitem{Chen2016}
\bibinfo{author}{Chen, T.} \& \bibinfo{author}{Guestrin, C.}
\newblock \bibinfo{title}{Xgboost: A scalable tree boosting system}.
\newblock In \emph{\bibinfo{booktitle}{Proceedings of the 22nd ACM SIGKDD International Conference on Knowledge Discovery and Data Mining}}, \bibinfo{pages}{785--794}, \doiprefix\url{10.1145/2939672.2939785} (\bibinfo{year}{2016}).

\bibitem{Lundberg2020}
\bibinfo{author}{Lundberg, S.~M.} \emph{et~al.}
\newblock \bibinfo{journal}{\bibinfo{title}{Explainable ai for trees: From local explanations to global understanding}}.
\newblock {\emph{\JournalTitle{Nature Machine Intelligence}}} \textbf{\bibinfo{volume}{2}}, \bibinfo{pages}{56--67}, \doiprefix\url{10.1038/s42256-019-0138-9} (\bibinfo{year}{2020}).

\bibitem{Choi2017}
\bibinfo{author}{Choi, E.}, \bibinfo{author}{Schuetz, A.}, \bibinfo{author}{Stewart, W.~F.} \& \bibinfo{author}{Sun, J.}
\newblock \bibinfo{journal}{\bibinfo{title}{Using recurrent neural network models for early detection of heart failure onset}}.
\newblock {\emph{\JournalTitle{Journal of the American Medical Informatics Association}}} \textbf{\bibinfo{volume}{24}}, \bibinfo{pages}{361--370}, \doiprefix\url{10.1093/jamia/ocw112} (\bibinfo{year}{2017}).

\bibitem{Katuwal2016}
\bibinfo{author}{Katuwal, G.~J.} \& \bibinfo{author}{Chen, J.}
\newblock \bibinfo{journal}{\bibinfo{title}{Feature selection and classification methods for imbalanced cancer datasets}}.
\newblock {\emph{\JournalTitle{PLoS ONE}}} \textbf{\bibinfo{volume}{11}}, \bibinfo{pages}{e0157853}, \doiprefix\url{10.1371/journal.pone.0157853} (\bibinfo{year}{2016}).

\bibitem{Luo2016}
\bibinfo{author}{Luo, Y.}, \bibinfo{author}{Szolovits, P.}, \bibinfo{author}{Dighe, A.} \& \bibinfo{author}{Baron, J.~M.}
\newblock \bibinfo{journal}{\bibinfo{title}{Using machine learning to predict laboratory test results}}.
\newblock {\emph{\JournalTitle{American Journal of Clinical Pathology}}} \textbf{\bibinfo{volume}{145}}, \bibinfo{pages}{778--788}, \doiprefix\url{10.1093/ajcp/aqw155} (\bibinfo{year}{2016}).

\bibitem{chawla2002smote}
\bibinfo{author}{Chawla, N.~V.}, \bibinfo{author}{Bowyer, K.~W.}, \bibinfo{author}{Hall, L.~O.} \& \bibinfo{author}{Kegelmeyer, W.~P.}
\newblock \bibinfo{journal}{\bibinfo{title}{Smote: synthetic minority over-sampling technique}}.
\newblock {\emph{\JournalTitle{Journal of artificial intelligence research}}} \textbf{\bibinfo{volume}{16}}, \bibinfo{pages}{321--357} (\bibinfo{year}{2002}).

\bibitem{elkan2001foundations}
\bibinfo{author}{Elkan, C.}
\newblock \bibinfo{title}{The foundations of cost-sensitive learning}.
\newblock In \emph{\bibinfo{booktitle}{International joint conference on artificial intelligence}}, vol.~\bibinfo{volume}{17}, \bibinfo{pages}{973--978} (\bibinfo{organization}{Lawrence Erlbaum Associates Ltd}, \bibinfo{year}{2001}).

\bibitem{lin2017focal}
\bibinfo{author}{Lin, T.-Y.}, \bibinfo{author}{Goyal, P.}, \bibinfo{author}{Girshick, R.}, \bibinfo{author}{He, K.} \& \bibinfo{author}{Doll{\'a}r, P.}
\newblock \bibinfo{title}{Focal loss for dense object detection}.
\newblock In \emph{\bibinfo{booktitle}{Proceedings of the IEEE international conference on computer vision}}, \bibinfo{pages}{2980--2988} (\bibinfo{year}{2017}).

\bibitem{Buda2018}
\bibinfo{author}{Buda, M.}, \bibinfo{author}{Maki, A.} \& \bibinfo{author}{Mazurowski, M.~A.}
\newblock \bibinfo{journal}{\bibinfo{title}{A systematic study of the class imbalance problem in convolutional neural networks}}.
\newblock {\emph{\JournalTitle{Neural Networks}}} \textbf{\bibinfo{volume}{106}}, \bibinfo{pages}{249--259}, \doiprefix\url{10.1016/j.neunet.2018.07.011} (\bibinfo{year}{2018}).

\bibitem{johnson2021mimic}
\bibinfo{author}{Johnson, A.} \emph{et~al.}
\newblock \bibinfo{journal}{\bibinfo{title}{Mimic-iv-ed}}.
\newblock {\emph{\JournalTitle{PhysioNet}}}  (\bibinfo{year}{2021}).

\bibitem{goldberger2000physiobank}
\bibinfo{author}{Goldberger, A.~L.} \emph{et~al.}
\newblock \bibinfo{journal}{\bibinfo{title}{Physiobank, physiotoolkit, and physionet: components of a new research resource for complex physiologic signals}}.
\newblock {\emph{\JournalTitle{circulation}}} \textbf{\bibinfo{volume}{101}}, \bibinfo{pages}{e215--e220} (\bibinfo{year}{2000}).

\bibitem{pollard2019eicu}
\bibinfo{author}{Pollard, T.} \emph{et~al.}
\newblock \bibinfo{journal}{\bibinfo{title}{eicu collaborative research database (version 2.0)}}.
\newblock {\emph{\JournalTitle{PhysioNet}}} \doiprefix\url{10.13026/C2WM1R} (\bibinfo{year}{2019}).

\bibitem{quinlan1986induction}
\bibinfo{author}{Quinlan, J.~R.}
\newblock \bibinfo{title}{Induction of decision trees}.
\newblock In \emph{\bibinfo{booktitle}{Machine Learning}}, vol.~\bibinfo{volume}{1}, \bibinfo{pages}{81--106} (\bibinfo{publisher}{Springer}, \bibinfo{year}{1986}).

\bibitem{breiman2001random}
\bibinfo{author}{Breiman, L.}
\newblock \emph{\bibinfo{title}{Random Forests}}, vol.~\bibinfo{volume}{45} (\bibinfo{publisher}{Machine Learning}, \bibinfo{year}{2001}).

\bibitem{chen2016xgboost}
\bibinfo{author}{Chen, T.} \& \bibinfo{author}{Guestrin, C.}
\newblock \bibinfo{title}{Xgboost: A scalable tree boosting system}.
\newblock In \emph{\bibinfo{booktitle}{Proceedings of the 22nd ACM SIGKDD International Conference on Knowledge Discovery and Data Mining}}, \bibinfo{pages}{785--794} (\bibinfo{organization}{ACM}, \bibinfo{year}{2016}).

\bibitem{ke2017lightgbm}
\bibinfo{author}{Ke, G.} \emph{et~al.}
\newblock \bibinfo{journal}{\bibinfo{title}{Lightgbm: A highly efficient gradient boosting decision tree}}.
\newblock {\emph{\JournalTitle{Advances in neural information processing systems}}} \textbf{\bibinfo{volume}{30}} (\bibinfo{year}{2017}).

\bibitem{prokhorenkova2018catboost}
\bibinfo{author}{Prokhorenkova, L.}, \bibinfo{author}{Gusev, G.}, \bibinfo{author}{Vorobev, A.}, \bibinfo{author}{Dorogush, A.~V.} \& \bibinfo{author}{Gulin, A.}
\newblock \bibinfo{journal}{\bibinfo{title}{Catboost: unbiased boosting with categorical features}}.
\newblock {\emph{\JournalTitle{Advances in neural information processing systems}}} \textbf{\bibinfo{volume}{31}} (\bibinfo{year}{2018}).

\bibitem{grinsztajn2022tree}
\bibinfo{author}{Grinsztajn, L.}, \bibinfo{author}{Oyallon, E.} \& \bibinfo{author}{Varoquaux, G.}
\newblock \bibinfo{journal}{\bibinfo{title}{Why do tree-based models still outperform deep learning on typical tabular data?}}
\newblock {\emph{\JournalTitle{Advances in neural information processing systems}}} \textbf{\bibinfo{volume}{35}}, \bibinfo{pages}{507--520} (\bibinfo{year}{2022}).

\bibitem{arik2021tabnet}
\bibinfo{author}{Arik, S.~{\"O}.} \& \bibinfo{author}{Pfister, T.}
\newblock \bibinfo{title}{Tabnet: Attentive interpretable tabular learning}.
\newblock In \emph{\bibinfo{booktitle}{Proceedings of the AAAI conference on artificial intelligence}}, vol.~\bibinfo{volume}{35}, \bibinfo{pages}{6679--6687} (\bibinfo{year}{2021}).

\bibitem{hollmann2022tabpfn}
\bibinfo{author}{Hollmann, N.}, \bibinfo{author}{M{\"u}ller, S.}, \bibinfo{author}{Eggensperger, K.} \& \bibinfo{author}{Hutter, F.}
\newblock \bibinfo{journal}{\bibinfo{title}{Tabpfn: A transformer that solves small tabular classification problems in a second}}.
\newblock {\emph{\JournalTitle{arXiv preprint arXiv:2207.01848}}}  (\bibinfo{year}{2022}).

\bibitem{grinsztajn2025tabpfn}
\bibinfo{author}{Grinsztajn, L.} \emph{et~al.}
\newblock \bibinfo{journal}{\bibinfo{title}{Tabpfn-2.5: Advancing the state of the art in tabular foundation models}}.
\newblock {\emph{\JournalTitle{arXiv preprint arXiv:2511.08667}}}  (\bibinfo{year}{2025}).

\bibitem{qu2025tabicl}
\bibinfo{author}{Qu, J.}, \bibinfo{author}{Holzm{\"u}ller, D.}, \bibinfo{author}{Varoquaux, G.} \& \bibinfo{author}{Le~Morvan, M.}
\newblock \bibinfo{title}{Tabicl: A tabular foundation model for in-context learning on large data}.
\newblock In \emph{\bibinfo{booktitle}{International Conference on Machine Learning}}, \bibinfo{pages}{50817--50847} (\bibinfo{organization}{PMLR}, \bibinfo{year}{2025}).

\bibitem{cui2019class}
\bibinfo{author}{Cui, Y.}, \bibinfo{author}{Jia, M.}, \bibinfo{author}{Lin, T.-Y.}, \bibinfo{author}{Song, Y.} \& \bibinfo{author}{Belongie, S.}
\newblock \bibinfo{title}{Class-balanced loss based on effective number of samples}.
\newblock In \emph{\bibinfo{booktitle}{Proceedings of the IEEE/CVF conference on computer vision and pattern recognition}}, \bibinfo{pages}{9268--9277} (\bibinfo{year}{2019}).

\bibitem{akiba2019optuna}
\bibinfo{author}{Akiba, T.}, \bibinfo{author}{Sano, S.}, \bibinfo{author}{Yanase, T.}, \bibinfo{author}{Ohta, T.} \& \bibinfo{author}{Koyama, M.}
\newblock \bibinfo{title}{Optuna: A next-generation hyperparameter optimization framework}.
\newblock In \emph{\bibinfo{booktitle}{Proceedings of the 25th ACM SIGKDD international conference on knowledge discovery \& data mining}}, \bibinfo{pages}{2623--2631} (\bibinfo{year}{2019}).

\bibitem{shwartz2021tabular}
\bibinfo{author}{Shwartz-Ziv, R.} \& \bibinfo{author}{Armon, A.}
\newblock \bibinfo{journal}{\bibinfo{title}{Tabular data: Deep learning is not all you need}}.
\newblock {\emph{\JournalTitle{Information Fusion}}} \textbf{\bibinfo{volume}{81}}, \bibinfo{pages}{84--90} (\bibinfo{year}{2022}).

\bibitem{lee2014validation}
\bibinfo{author}{Lee, H.}, \bibinfo{author}{Shon, Y.-J.}, \bibinfo{author}{Kim, H.}, \bibinfo{author}{Paik, H.} \& \bibinfo{author}{Park, H.-P.}
\newblock \bibinfo{journal}{\bibinfo{title}{Validation of the apache iv model and its comparison with the apache ii, saps 3, and korean saps 3 models for the prediction of hospital mortality in a korean surgical intensive care unit}}.
\newblock {\emph{\JournalTitle{Korean journal of anesthesiology}}} \textbf{\bibinfo{volume}{67}}, \bibinfo{pages}{115} (\bibinfo{year}{2014}).

\bibitem{badrinath2018comparison}
\bibinfo{author}{Badrinath, K.} \emph{et~al.}
\newblock \bibinfo{journal}{\bibinfo{title}{Comparison of various severity assessment scoring systems in patients with sepsis in a tertiary care teaching hospital}}.
\newblock {\emph{\JournalTitle{Indian journal of critical care medicine: peer-reviewed, official publication of Indian Society of Critical Care Medicine}}} \textbf{\bibinfo{volume}{22}}, \bibinfo{pages}{842} (\bibinfo{year}{2018}).

\bibitem{minne2008evaluation}
\bibinfo{author}{Minne, L.}, \bibinfo{author}{Abu-Hanna, A.} \& \bibinfo{author}{de~Jonge, E.}
\newblock \bibinfo{journal}{\bibinfo{title}{Evaluation of sofa-based models for predicting mortality in the icu: A systematic review}}.
\newblock {\emph{\JournalTitle{Critical care}}} \textbf{\bibinfo{volume}{12}}, \bibinfo{pages}{R161} (\bibinfo{year}{2008}).

\bibitem{bosch2022predictive}
\bibinfo{author}{Bosch, N.~A.}, \bibinfo{author}{Law, A.~C.}, \bibinfo{author}{Rucci, J.~M.}, \bibinfo{author}{Peterson, D.} \& \bibinfo{author}{Walkey, A.~J.}
\newblock \bibinfo{journal}{\bibinfo{title}{Predictive validity of the sequential organ failure assessment score versus claims-based scores among critically ill patients}}.
\newblock {\emph{\JournalTitle{Annals of the American Thoracic Society}}} \textbf{\bibinfo{volume}{19}}, \bibinfo{pages}{1072--1076} (\bibinfo{year}{2022}).

\bibitem{asmarawati2022predictive}
\bibinfo{author}{Asmarawati, T.~P.} \emph{et~al.}
\newblock \bibinfo{journal}{\bibinfo{title}{Predictive value of sequential organ failure assessment, quick sequential organ failure assessment, acute physiology and chronic health evaluation ii, and new early warning signs scores estimate mortality of covid-19 patients requiring intensive care unit}}.
\newblock {\emph{\JournalTitle{Indian Journal of Critical Care Medicine: Peer-reviewed, Official Publication of Indian Society of Critical Care Medicine}}} \textbf{\bibinfo{volume}{26}}, \bibinfo{pages}{464} (\bibinfo{year}{2022}).

\bibitem{ancker2017effects}
\bibinfo{author}{Ancker, J.~S.} \emph{et~al.}
\newblock \bibinfo{journal}{\bibinfo{title}{Effects of workload, work complexity, and repeated alerts on alert fatigue in a clinical decision support system}}.
\newblock {\emph{\JournalTitle{BMC medical informatics and decision making}}} \textbf{\bibinfo{volume}{17}}, \bibinfo{pages}{36} (\bibinfo{year}{2017}).

\bibitem{chaparro2022clinical}
\bibinfo{author}{Chaparro, J.~D.} \emph{et~al.}
\newblock \bibinfo{journal}{\bibinfo{title}{Clinical decision support stewardship: best practices and techniques to monitor and improve interruptive alerts}}.
\newblock {\emph{\JournalTitle{Applied Clinical Informatics}}} \textbf{\bibinfo{volume}{13}}, \bibinfo{pages}{560--568} (\bibinfo{year}{2022}).

\bibitem{incremona2025differentiable}
\bibinfo{author}{Incremona, A.}, \bibinfo{author}{Pozzi, A.}, \bibinfo{author}{Guiscardi, A.} \& \bibinfo{author}{Tessera, D.}
\newblock \bibinfo{journal}{\bibinfo{title}{A differentiable and uncertainty-aware mutual information regularizer for bias mitigation}}.
\newblock {\emph{\JournalTitle{Neurocomputing}}} \bibinfo{pages}{132498} (\bibinfo{year}{2025}).

\bibitem{donini2018empirical}
\bibinfo{author}{Donini, M.}, \bibinfo{author}{Oneto, L.}, \bibinfo{author}{Ben-David, S.}, \bibinfo{author}{Shawe-Taylor, J.~S.} \& \bibinfo{author}{Pontil, M.}
\newblock \bibinfo{journal}{\bibinfo{title}{Empirical risk minimization under fairness constraints}}.
\newblock {\emph{\JournalTitle{Advances in neural information processing systems}}} \textbf{\bibinfo{volume}{31}} (\bibinfo{year}{2018}).

\bibitem{oneto2020general}
\bibinfo{author}{Oneto, L.}, \bibinfo{author}{Donini, M.} \& \bibinfo{author}{Pontil, M.}
\newblock \bibinfo{title}{General fair empirical risk minimization}.
\newblock In \emph{\bibinfo{booktitle}{2020 International Joint Conference on Neural Networks (IJCNN)}}, \bibinfo{pages}{1--8} (\bibinfo{organization}{IEEE}, \bibinfo{year}{2020}).

\bibitem{abramoff2023considerations}
\bibinfo{author}{Abr{\`a}moff, M.~D.} \emph{et~al.}
\newblock \bibinfo{journal}{\bibinfo{title}{Considerations for addressing bias in artificial intelligence for health equity}}.
\newblock {\emph{\JournalTitle{NPJ digital medicine}}} \textbf{\bibinfo{volume}{6}}, \bibinfo{pages}{170} (\bibinfo{year}{2023}).

\bibitem{sikstrom2022conceptualising}
\bibinfo{author}{Sikstrom, L.} \emph{et~al.}
\newblock \bibinfo{journal}{\bibinfo{title}{Conceptualising fairness: three pillars for medical algorithms and health equity}}.
\newblock {\emph{\JournalTitle{BMJ health \& care informatics}}} \textbf{\bibinfo{volume}{29}}, \bibinfo{pages}{e100459} (\bibinfo{year}{2022}).

\bibitem{rajkomar2018scalable}
\bibinfo{author}{Rajkomar, A.} \emph{et~al.}
\newblock \bibinfo{journal}{\bibinfo{title}{Scalable and accurate deep learning with electronic health records}}.
\newblock {\emph{\JournalTitle{NPJ digital medicine}}} \textbf{\bibinfo{volume}{1}}, \bibinfo{pages}{18} (\bibinfo{year}{2018}).

\bibitem{garg2025real}
\bibinfo{author}{Garg, A.} \emph{et~al.}
\newblock \bibinfo{journal}{\bibinfo{title}{Real-tabpfn: Improving tabular foundation models via continued pre-training with real-world data}}.
\newblock {\emph{\JournalTitle{arXiv preprint arXiv:2507.03971}}}  (\bibinfo{year}{2025}).

\bibitem{ma2024tabdpt}
\bibinfo{author}{Ma, J.} \emph{et~al.}
\newblock \bibinfo{journal}{\bibinfo{title}{Tabdpt: Scaling tabular foundation models on real data}}.
\newblock {\emph{\JournalTitle{arXiv preprint arXiv:2410.18164}}}  (\bibinfo{year}{2024}).

\bibitem{johnson2024mimic}
\bibinfo{author}{Johnson, A.}, \bibinfo{author}{Pollard, T.}, \bibinfo{author}{Mark, R.}, \bibinfo{author}{Berkowitz, S.} \& \bibinfo{author}{Horng, S.}
\newblock \bibinfo{journal}{\bibinfo{title}{Mimic-cxr database}}.
\newblock {\emph{\JournalTitle{PhysioNet10}}} \textbf{\bibinfo{volume}{13026}}, \bibinfo{pages}{C2JT1Q} (\bibinfo{year}{2024}).

\end{thebibliography}

\section*{Funding Statement}
This research was supported in part by the National Research Foundation of South Africa (Ref No. CSRP23040990793).

\section*{Author contributions statement}
Y.B. conceived and designed the study, curated and preprocessed the datasets, implemented the machine learning models, performed the experiments, analyzed the results, and drafted the manuscript. M.A. contributed to the methodological design, supervised the work, assisted with interpretation of the findings, and provided critical revisions to the manuscript. All authors read and approved the final manuscript.

\section*{Competing Interests}
All authors declare no financial or non-financial competing interests.   

\section*{Data Availability}
The datasets analyzed during the current study are freely available and hosted on PhysioNet. 
The MIMIC-IV-ED (v2.2), MIMIC-CXR (v2.1.0), and eICU Collaborative Research Database (v2.0) are available at 
\url{https://physionet.org/content/mimic-iv-ed/2.2/}, 
\url{https://physionet.org/content/mimic-cxr/2.1.0/}, 
and \url{https://physionet.org/content/eicu-crd/2.0/}, respectively. 
Access to these databases requires credentialing and completion of a data use agreement through PhysioNet. No new datasets were generated in this study.

\section*{Code Availability}
The code developed for this study is publicly available on GitHub at \url{https://github.com/yusufbrima/tabresnet}.

\appendix

\section{Data Preprocessing and Feature Engineering}
\label{appendix:preprocessing}
\subsection{MIMIC-IV-ED}
\label{appendix:preprocessing_mimic}
For the MIMIC-IV-ED cohort, we extracted structured electronic health records encompassing patient stays, triage assessments, diagnoses, and vital sign measurements. The initial cohort definition leveraged the MIMIC-CXR~\cite{johnson2024mimic} core table to identify eligible patients and establish potential linkages with radiographic data. In this study, however, we focused our analysis to structured tabular information from the emergency department (ED), while noting that the multimodal potential of this resource remains valuable for future investigations.

Within this structured dataset, diagnostic information was standardized by reducing ICD codes to the three-character level, thereby capturing broader diagnostic categories in a clinically interpretable manner. For each ED stay, the first recorded diagnosis was designated as the primary label, and patient disposition at discharge was harmonized into four outcome categories (admitted, transferred, discharged, deceased) through a standardized mapping procedure.

In addition to diagnostic and outcome information, vital sign measurements were aggregated at the stay level to summarize temperature, heart rate, respiratory rate, oxygen saturation, and systolic/diastolic blood pressure. This approach reduced redundancy from repeated recordings while retaining clinically relevant signals. Demographic and ED-related variables (e.g., age, sex, race, mode of arrival, and acuity) were also preserved as predictors, providing contextual information about patient presentation.

To maintain data integrity, records were restricted to those with valid ICD codes and complete linkages between ED stays and diagnoses, and duplicate identifiers were removed to avoid overrepresentation. The resulting curated dataset consisted of identifiers, target variables (diagnoses, grouped ICD codes, discharge disposition), demographic attributes, ED-related features, and aggregated vital signs. After preprocessing, this streamlined and clinically interpretable EHR cohort provided a robust foundation for predictive modeling, while its multimodal structure points to future opportunities for integrating imaging with EHR data.

\subsection{eICU Collaborative Research Database}
\label{appendix:preprocessing_eicu}
For this dataset, we constructed the cohort with careful attention to temporal leakage prevention. Core patient information, admission characteristics, diagnostic data, laboratory measurements, and periodic vital sign recordings were extracted. To ensure that models relied only on information available at clinically relevant decision points, we restricted measurements to the first 24 hours of ICU admission. Vital signs and laboratory results were filtered using time offsets, preventing any leakage from future observations.

Building on this temporal window, vital sign features were aggregated at the patient-stay level, with summary statistics (mean, standard deviation, minimum, maximum, and count) computed for variables such as temperature, oxygen saturation, heart rate, respiration rate, and blood pressure. Laboratory measurements were similarly constrained to the 20 most frequently ordered tests and summarized with equivalent statistics. Admission diagnoses were limited to those recorded at or before admission, and these were represented both as concatenated diagnosis strings and as diagnosis counts.

From these features, we derived a set of clinically meaningful target variables. Length of stay was categorized into five ordinal classes, ranging from very short (<24 h) to prolonged (>4 weeks). Clinical severity was operationalized via a composite score integrating indicators of organ support (e.g., ventilation, dialysis), circulatory compromise, neurological status (Glasgow Coma Scale), and metabolic derangement. Discharge disposition was harmonized into interpretable categories, including home/hospice, extended care facilities, ICU transfers, and death. Resource utilization was characterized by combining length of stay with high-intensity interventions, stratifying patients into four utilization tiers.

The final merged dataset brought together demographic variables, severity scores, interventions, aggregated vital signs, laboratory summaries, and diagnostic features alongside the constructed target variables. Data quality checks were applied to remove potential leakage columns (e.g., hospital discharge status), and records with missing target labels were excluded. Categorical variables, including gender, ethnicity, and diagnosis strings, were retained in their raw form to allow flexible encoding strategies. To facilitate downstream modeling, the processed datasets were saved in modular format, separating features, targets, and the complete merged dataset for streamlined machine learning pipelines.

\subsection{Feature Normalization and Handling Missing Values}
To ensure comparability across heterogeneous clinical variables, continuous features were standardized to zero mean and unit variance prior to model training. Categorical features, including demographic indicators and diagnostic codes, were transformed via one-hot encoding to preserve discrete class membership without imposing ordinal assumptions. Missing values were addressed through simple but robust imputation strategies: continuous variables were imputed with the median, whereas categorical variables were imputed with the mode. To maintain data integrity, features with more than 50\% missing data were excluded from the analytic dataset. This procedure balances the trade-off between retaining as much information as possible and avoiding the introduction of excessive noise from sparsely observed features.

\subsection{Train–Validation–Test Splits}
For both the MIMIC-IV-ED and eICU cohorts, we partitioned each into training, validation, and test sets with proportions of 60\%, 20\%, and 20\%, respectively. Stratified sampling was applied to preserve the distribution of outcome classes across all subsets, thereby ensuring that rare but clinically important categories remained represented in both model development and evaluation. This strategy facilitated reliable performance assessment while guarding against biased estimates due to class imbalance.
\subsection{Hyperparameter Optimization}
\label{appendix:hyperparameters}
This section summarizes the hyperparameter search spaces, optimization procedures, and evaluation strategies applied across all the models. Optimization was performed with Optuna via the tree-structured Parzen estimator (TPE; seed=42) and the MedianPruner to terminate unpromising trials. Each model–dataset pair was tuned over 100 trials, and performance was consistently measured by the average F1 score on the validation set.

\subsection{Traditional Machine Learning Models}
Decision trees were tuned over maximum depth (2–32), minimum samples per split (2–50), minimum samples per leaf (1–20), and splitting criterion (gini or entropy). Random forests varied in number of estimators (100–1000), depth (3–25), splitting parameters as above, and maximum features (sqrt, log2, or fixed fractions). XGBoost models were optimized for estimators (200–1200), learning rate (0.01–0.3), depth (3–12), sub-sampling (0.6–1.0), column sampling (0.5–1.0), and regularization terms ($\alpha$ and $\lambda$, 0–5).

\subsection{Deep Learning Models}
TabNet was tuned over learning rate ($1 \times 10^{-6}$–$1 \times 10^{-1}$), weight decay ($1 \times 10^{-7}$–$1 \times 10^{-2}$), and batch size (32–1024). 
\subsubsection{Evaluation and Convergence}
For traditional models, hyperparameters were selected by training on the training split and evaluating on the validation split. The Deep learning models followed the same scheme with early stopping: training halted after 15 epochs without validation improvement for TabNet, with the best checkpoint restored. A ReduceLROnPlateau scheduler reduced learning rates by a factor of 0.5 after three stagnant epochs.  
\subsection{Class Imbalance Handling}
To address class imbalance, weighting strategies were incorporated during optimization. For traditional models, class or sample weights were applied directly through implementation parameters. For deep learning, class weights were embedded in the loss function. We compared inverse frequency, median frequency, and effective number of samples (with $\beta=0.9999$), as well as unweighted baselines.

\section{MIMIC-IV-ED Results}
\label{appendix:extended_results}
\subsubsection{Training Performance Across Filter Sizes}
\label{mimic_training_filter_size_plots}
\begin{figure}[!ht]
    \centering
    \includegraphics[width=0.90\linewidth]{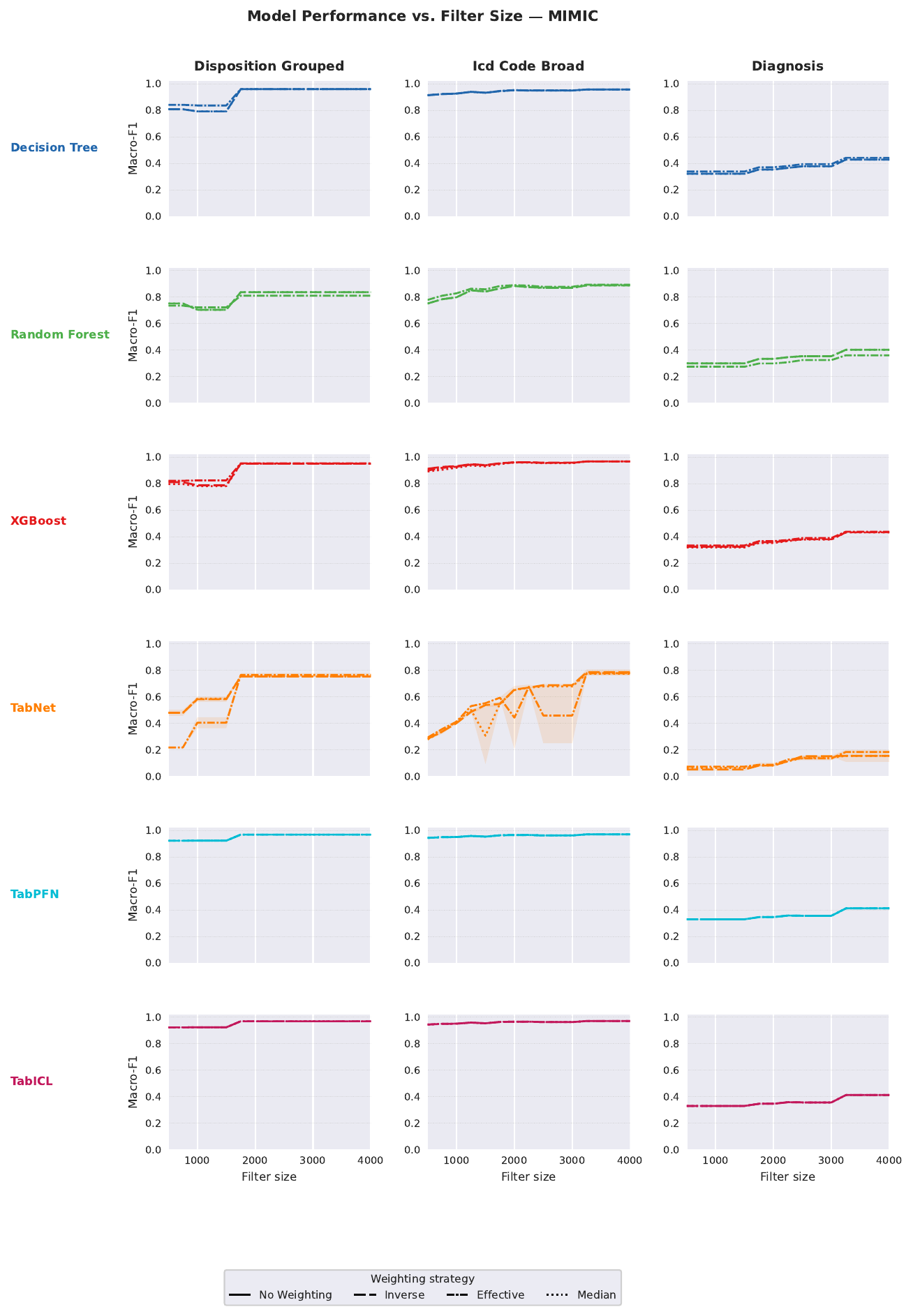}
    
    \caption{\textbf{\textbf{Model performance  across different filter sizes on the MIMIC-iV-ED database}}}
    \label{fig:performance_vs_filter_size_mimic}
\end{figure}

Figure~\ref{fig:performance_vs_filter_size_mimic} provides extended analysis of training time scaling with dataset size for the MIMIC-IV-ED dataset.

\subsubsection{Training Time Scaling Plots}
\label{mimic_training_time_plots}
\begin{figure}[!ht]
    \centering
    \includegraphics[width=.80\linewidth]{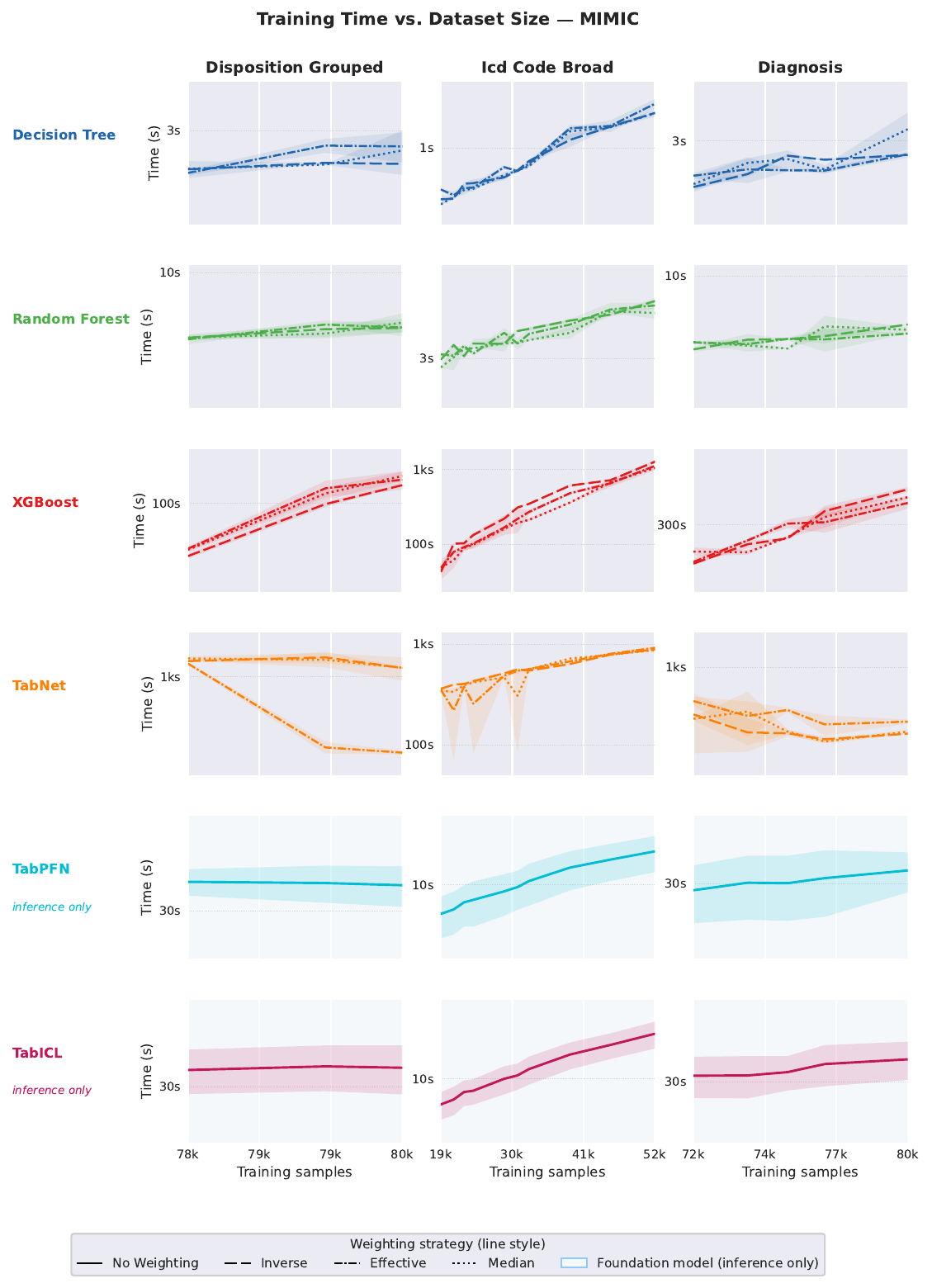}
    
    \caption{\textbf{Computational scaling across model architectures.} Training time as a function of dataset size for different prediction tasks. Each panel shows the training time (seconds, log scale) versus the total number of training samples for a specific target variable. The results are reported across all classifiers and class weighting strategies.}
    \label{fig:model_time_performance_mimic_all_targets}
\end{figure}

Figure~\ref{fig:model_time_performance_mimic_all_targets} provides extended analysis of training time scaling with dataset size for the MIMIC-IV-ED dataset. Each panel corresponds to one of the three prediction tasks: disposition outcomes, ICD code categories, and primary diagnosis. The training time is plotted on a logarithmic scale against the number of training samples, with curves shown for all classifiers and class weighting strategies. These plots complement the rank-based comparisons presented in the main text by highlighting absolute training time differences and scaling behavior across models. In particular, they illustrate the widening efficiency gap between tree-based ensembles and attention-based deep learning models as the dataset size increases.

\section{eICU-CRD Results}
\label{appendix:eicu_results}
Here, we present extended results for the eICU-CRD dataset, complementing the main text findings by providing detailed analyses of class imbalance metrics, classifier performance, and training time behavior across diverse prediction tasks. The figures included here are intended to provide additional depth, showing how different models and weighting strategies perform across length of stay, severity, discharge disposition, and resource utilization outcomes.

\subsection{Classifier Performance Comparison}
\label{eicu_model_performance}
\begin{figure}[!ht]
    \centering
    \includegraphics[width=\textwidth]{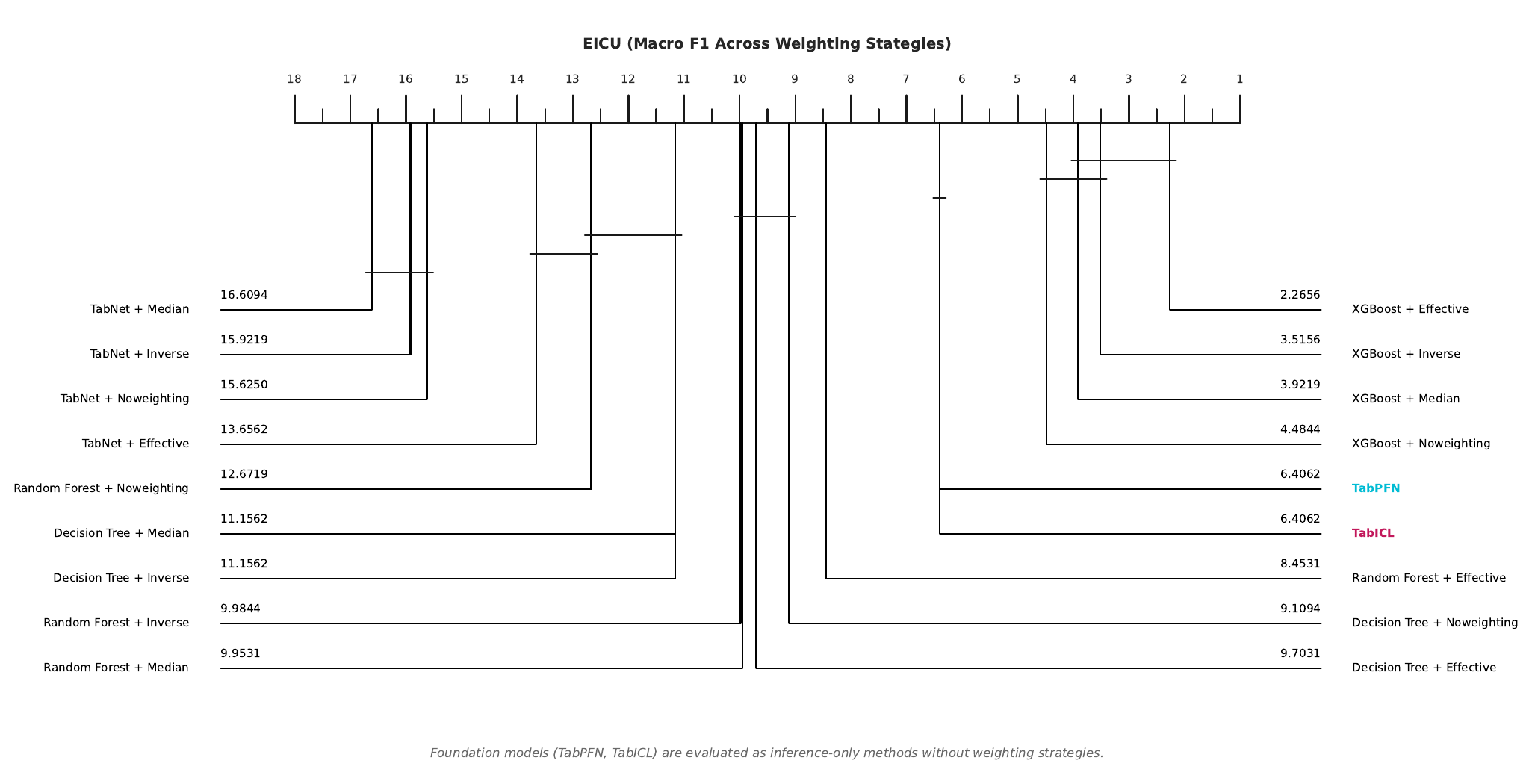}
    \caption{\textbf{Classifier performance rankings across experimental conditions.} Critical difference diagram showing the average ranks of 18 classifiers on the eICU-CRD dataset on the basis of Macro F1 scores across experimental blocks. Lower ranks indicate better predictive performance. The classifiers connected by a horizontal bar are not significantly different according to Wilcoxon signed-rank tests with Holm correction.}
    \label{fig:critical_difference_diagram_eicu_all_classifiers}
\end{figure}

We compared the predictive performance (Macro F1 score) of 18 classifier configurations across multiple experimental blocks, with each block corresponding to a unique combination of a target variable and a training set size. The 18 configurations span six model families, where four trainable models (Decision Tree, Random Forest, XGBoost, and TabNet) were each evaluated under four class weighting strategies, and two inference-only foundation models (TabPFN v2.6 and TabICL) were evaluated in their pretrained regimes without weighting. A Friedman test indicated significant differences among classifier configurations \RW{($\chi^2(17, N = 32) = 374.83$, $p = 3.34 \times 10^{-69}$).} Post-hoc pairwise comparisons with Wilcoxon signed-rank tests and Holm correction (Figure~\ref{fig:critical_difference_diagram_eicu_all_classifiers}) revealed a consistent pattern in which XGBoost variants achieved the lowest average ranks, reflecting comparatively stronger predictive performance on this dataset, followed by TabICL and TabPFN v2.6. Random Forest and Decision Tree variants occupied intermediate positions, performing competitively in several tasks but without consistently matching gradient-boosting performance on this dataset. TabNet variants obtained the highest average ranks, indicating comparatively weaker performance under the eICU-CRD conditions evaluated here.

\RW{These cross-dataset differences, particularly the more pronounced ranking of foundation-based models on eICU-CRD relative to MIMIC-IV-ED, suggest that the relative advantage of inference-based tabular models may be sensitive to dataset characteristics such as scale, label structure, and distributional properties.}

Across all tasks, performance degradation curves were broadly consistent across imbalance measures: higher IR and CVCF values, or lower NECD values (approaching 0 from 1), were associated with declines in Macro F1, though the magnitude of degradation varied by model family and prediction target.

\begin{figure}[!ht] 
    \centering
    \includegraphics[width=1.0\textwidth]{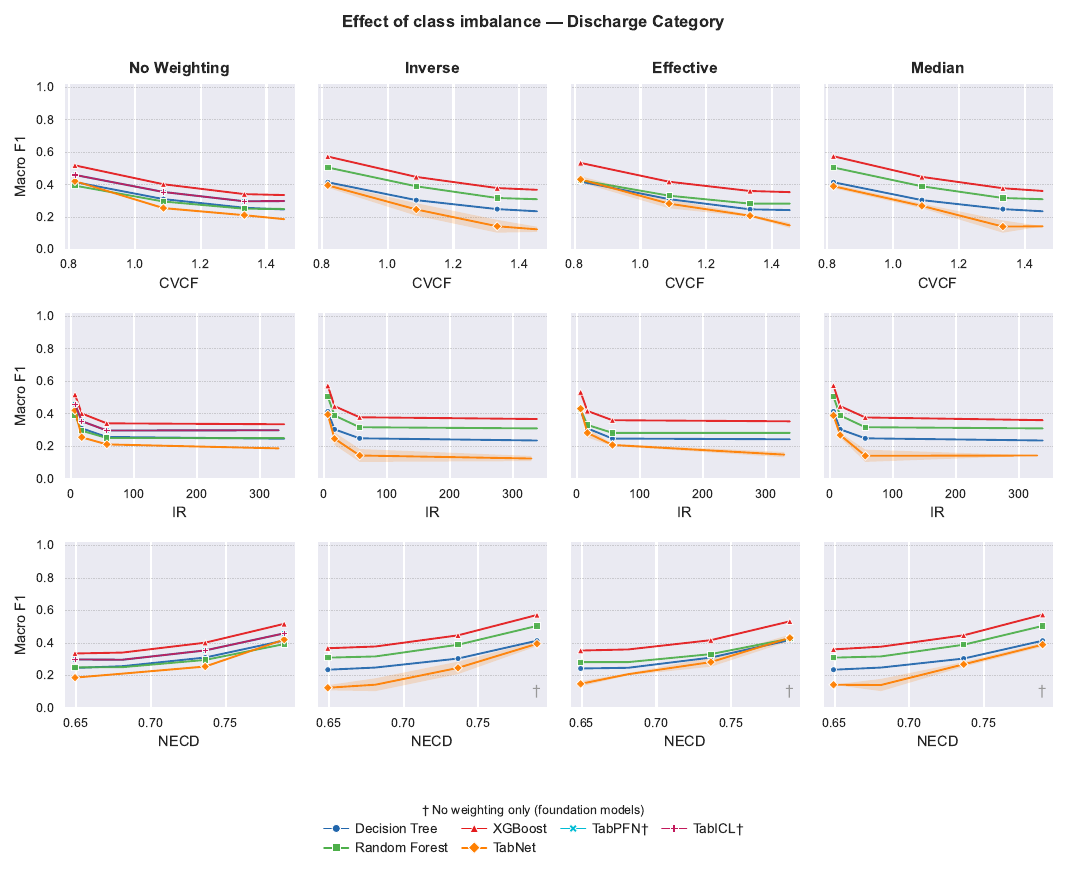} 
    \caption{\textbf{Impact of class imbalance on discharge disposition prediction.} 
    Macro F1 outcomes for 6 model families under four weighting strategies. Performance trends are shown with respect to three imbalance measures. Tree-based ensemble approaches, particularly XGBoost, were generally among the more robust models across imbalance levels, whereas deep models tended to show steeper declines in some settings. CVCF trends were broadly consistent with those from IR and NECD.}
    \label{fig:class_imbalance_effect_eicu_discharge_category}
\end{figure}

\begin{figure}[!ht] 
    \centering
    \includegraphics[width=1.0\textwidth]{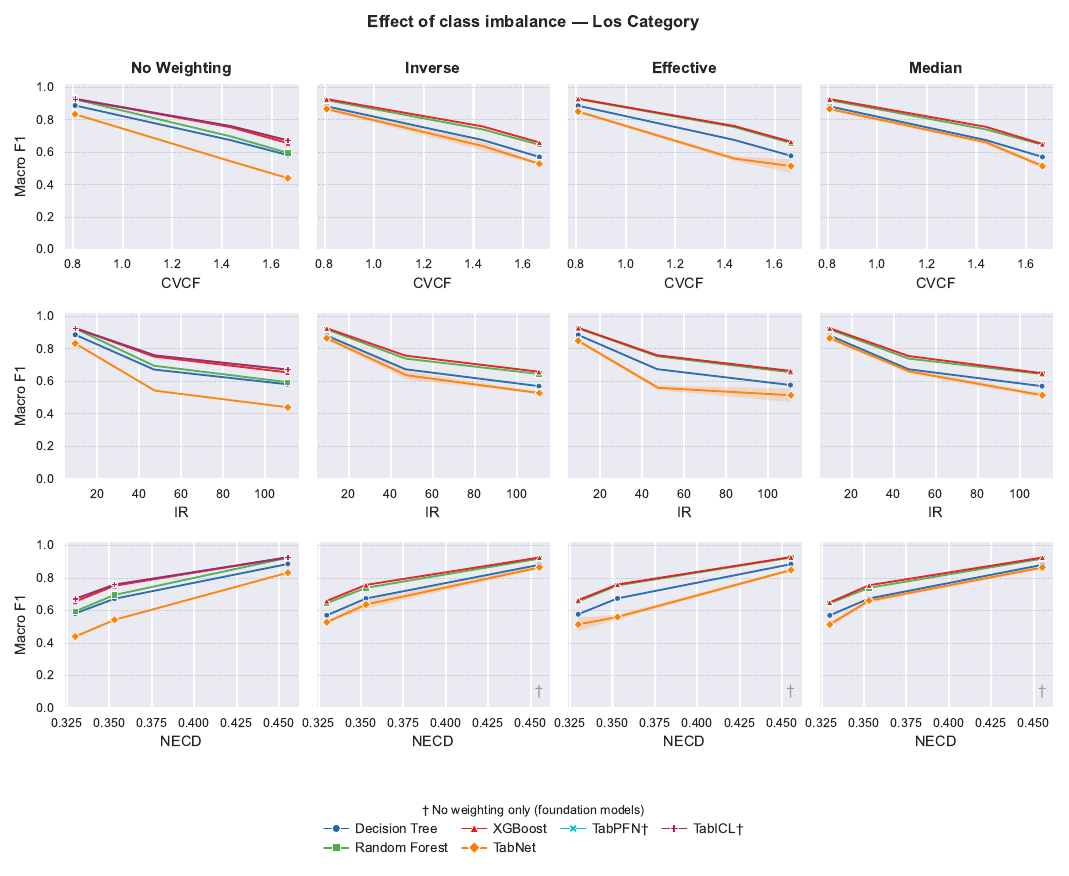} 
    \caption{\textbf{Class imbalance and length-of-stay prediction.} 
    Macro F1 trajectories for the same model families across different weighting strategies, plotted against CVCF, IR, and NECD. Although performance generally decreased as imbalance increased, the magnitude of the decline varied by model family and weighting scheme.}
    \label{fig:class_imbalance_effect_eicu_los_category}
\end{figure}

\begin{figure}[!ht] 
    \centering
    \includegraphics[width=1.0\textwidth]{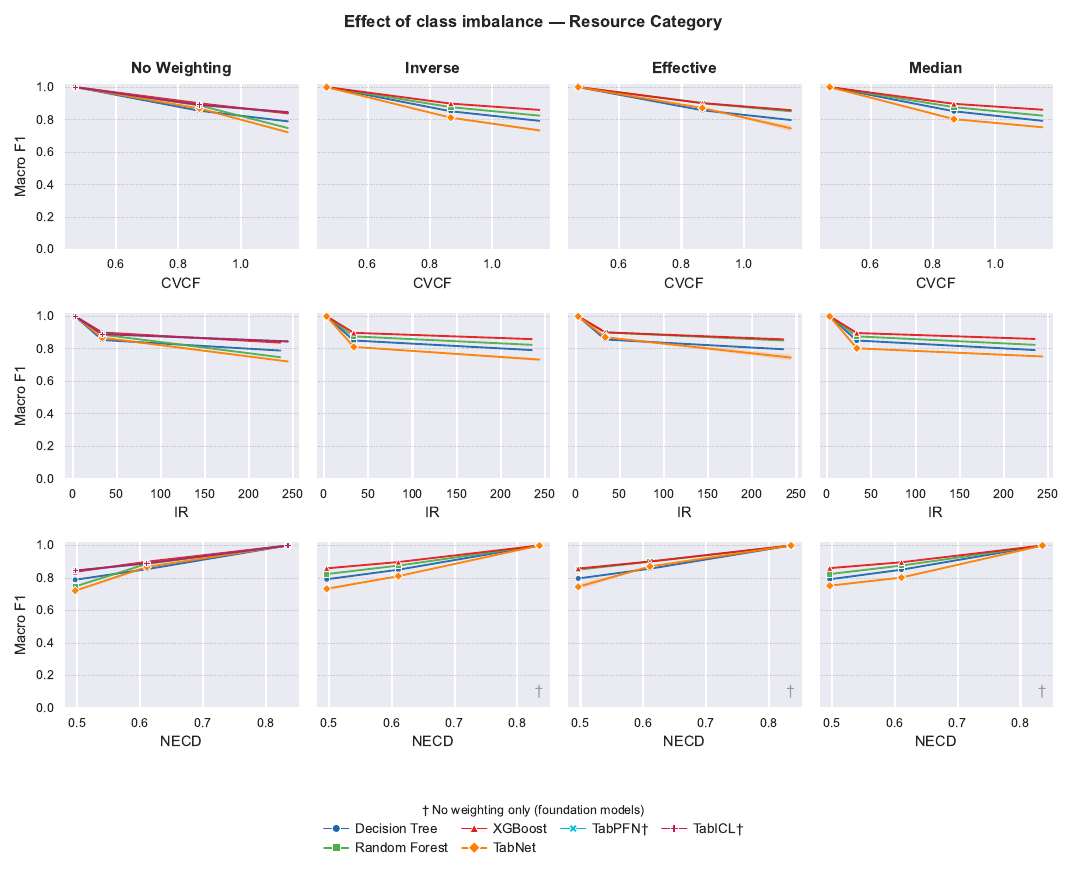} 
    \caption{\textbf{Influence of imbalance on resource utilization prediction.} 
    Macro F1 values for seven model families using four weighting schemes. Results are tracked across three imbalance metrics. Most models showed gradual declines in performance, with deep learning methods appearing more sensitive to skew in several settings, whereas ensemble methods tended to remain more stable. CVCF was somewhat more variable but remained directionally consistent with IR and NECD, and NECD decreased as imbalance increased.}
    \label{fig:class_imbalance_effect_eicu_resource_category}
\end{figure}

\begin{figure}[!ht] 
    \centering
    \includegraphics[width=1.0\textwidth]{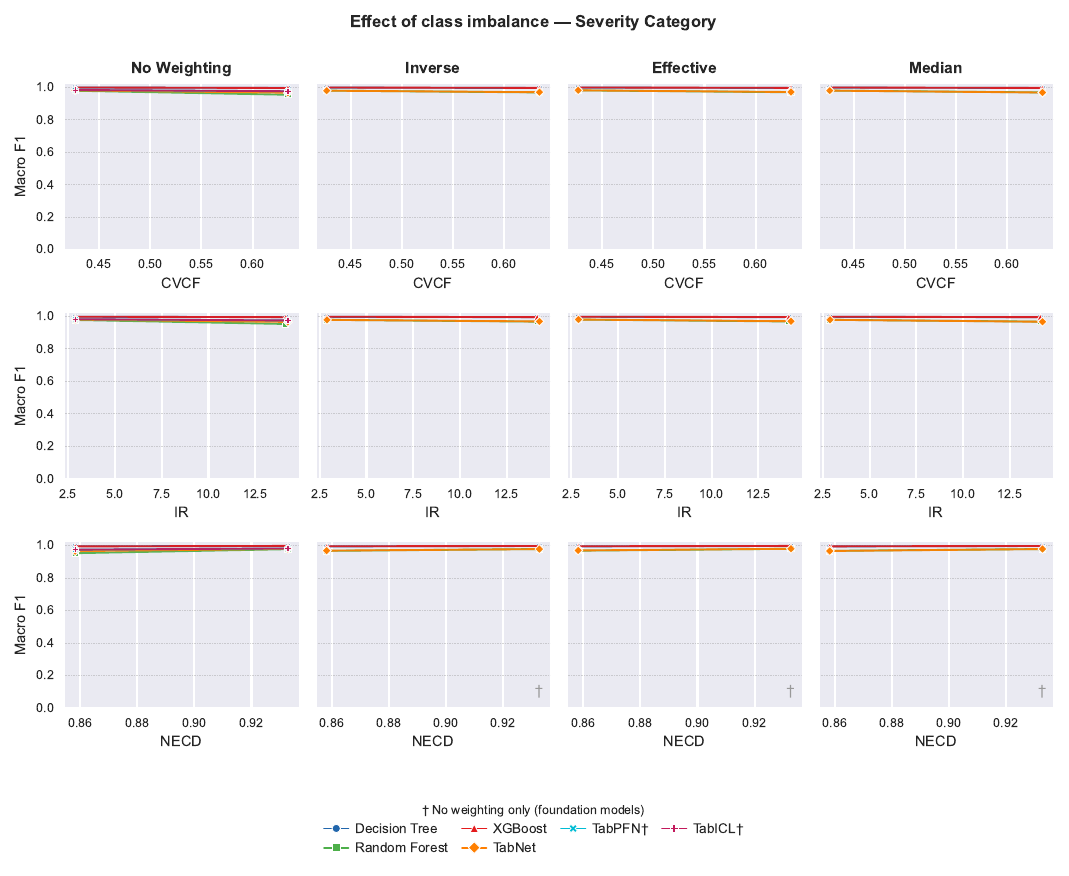} 
    \caption{\textbf{Performance under class imbalance for severity prediction.} 
    Macro F1 results comparing all model families and weighting strategies against CVCF, IR, and NECD. Severity classification appeared somewhat less sensitive to imbalance than some of the other tasks. The three imbalance metrics yielded broadly similar performance curves, with NECD decreasing monotonically as imbalance increased.}
    \label{fig:class_imbalance_effect_eicu_severity_category}
\end{figure}

Extended results for individual prediction tasks are shown in Figures~\ref{fig:class_imbalance_effect_eicu_discharge_category}--\ref{fig:class_imbalance_effect_eicu_severity_category}. These task-level analyses show that performance generally declined as imbalance became more pronounced, although the size of the decline varied across outcomes, model families, and weighting strategies. Tree-based ensembles, particularly XGBoost, were often among the stronger performers on eICU-CRD, whereas TabNet and TabResNet tended to be more sensitive to imbalance. TabPFN v2.6 and TabICL were competitive in several settings, but their relative standing was not uniform across tasks. Across panels, CVCF followed broadly similar trends to IR and NECD, while NECD decreased monotonically as imbalance increased.

\subsection{Training Time Comparison}
\label{eicu_training_time}
\begin{figure}[!ht]
    \centering
    \includegraphics[width=\textwidth]{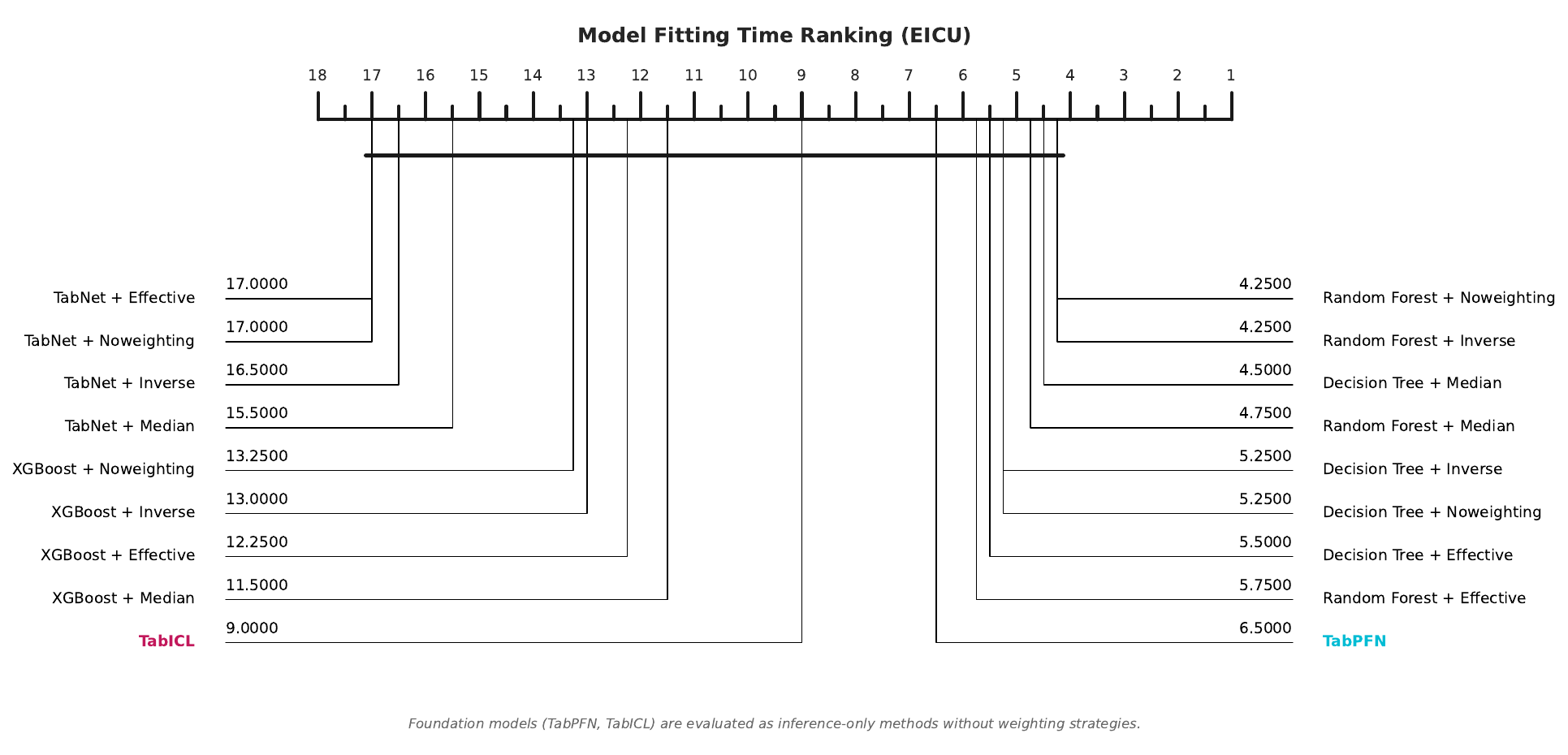}
    \caption{\textbf{Critical difference diagram for classifier training times.} 
    Critical difference diagram of the average ranks of 18 classifiers on this dataset, based on training times across experimental blocks. Lower ranks indicate faster training. Horizontal bars connect classifiers that are not significantly different under Wilcoxon signed-rank tests with Holm correction.}
\label{fig:critical_difference_diagram_eicu_training_time_with_samples}
\end{figure}

We compared the training times of 18 classifier configurations across 32 experimental blocks, where each block corresponds to a unique combination of a target variable and a training set size. A Friedman test revealed a statistically significant difference in training times among the classifiers \RW{($\chi^2(17, N = 32) = 58.04$, $p = 2.21 \times 10^{-6}$).} Post-hoc pairwise comparisons using Wilcoxon signed-rank tests with Holm correction (Figure~\ref{fig:critical_difference_diagram_eicu_training_time_with_samples}) indicated clear differences in computational cost across model families. Classical tree-based methods were generally the most efficient, while TabNet was consistently the slowest.

\RW{The remaining families, including TabPFN v2.6 and TabICL, occupied intermediate positions in terms of measured runtime. However, it is important to note that both TabPFN and TabICL follow an inference-based (in-context learning) paradigm and do not involve conventional gradient-based training on downstream tasks. Therefore, their reported timing reflects the cost of in-context conditioning and inference over the provided support set rather than full model optimization. As such, their runtime should be interpreted as task-level adaptation and inference cost rather than training cost in the classical sense.}

These results suggest that computational cost is driven more by model family and learning paradigm than by the particular class weighting scheme used.

\subsection{Training Performance Across Filter Sizes}
\label{eicu:training_filter_size_curves}
\begin{figure}[!ht]
    \centering
    \includegraphics[width=.950\linewidth]{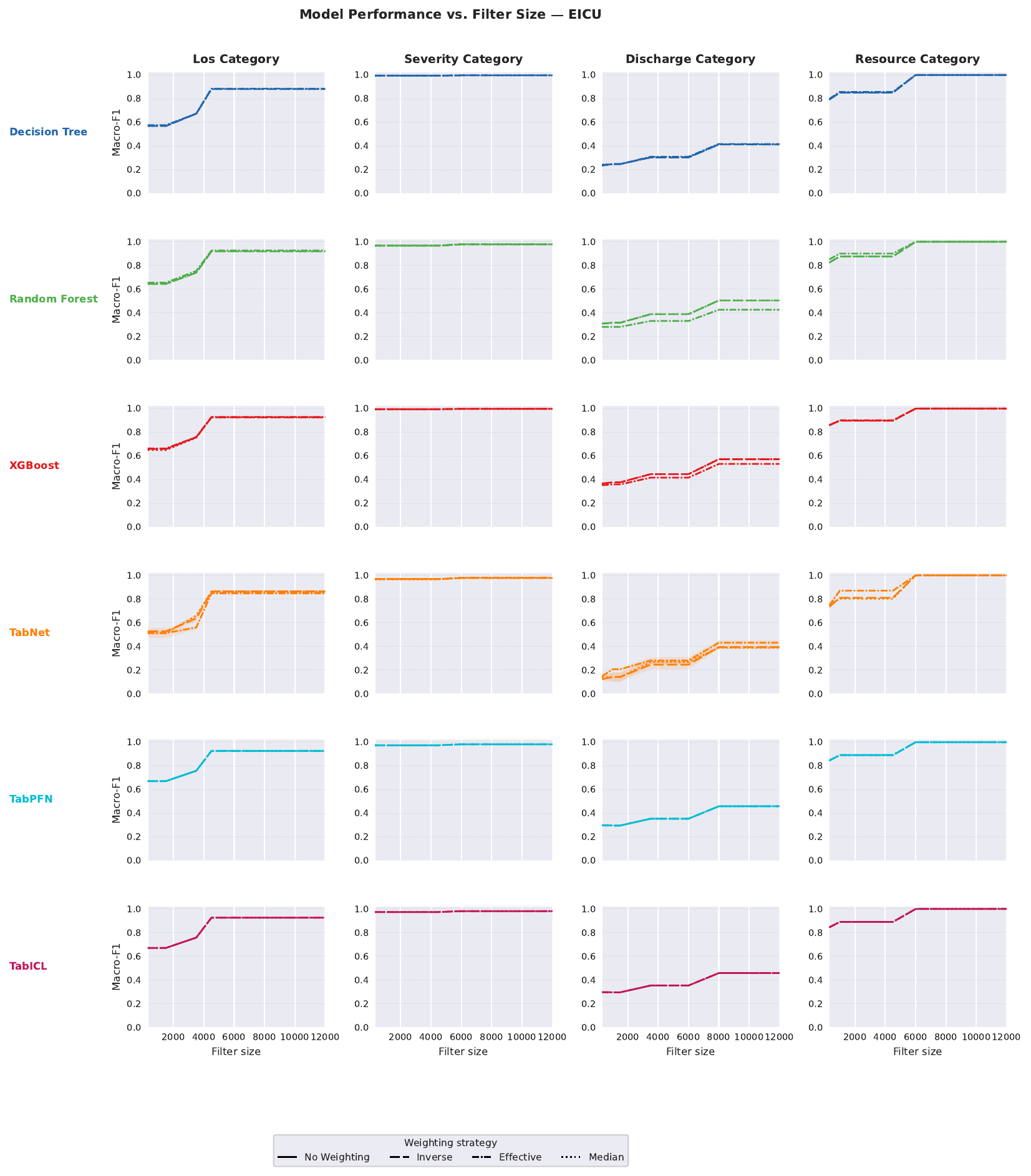}
    \caption{\textbf{Model performance across the different filter sizes on the eICU database.}}
    \label{fig:performance_vs_filter_size_eicu}
\end{figure}

Figure~\ref{fig:performance_vs_filter_size_eicu} presents extended analyses of training time scaling with dataset size across the seven prediction tasks in the eICU-CRD dataset. 

\subsection{Training Time Scaling Plots}
\label{eicu:training_time_curves}
\begin{figure}[!htp]
    \centering
    \includegraphics[width=.850\linewidth]{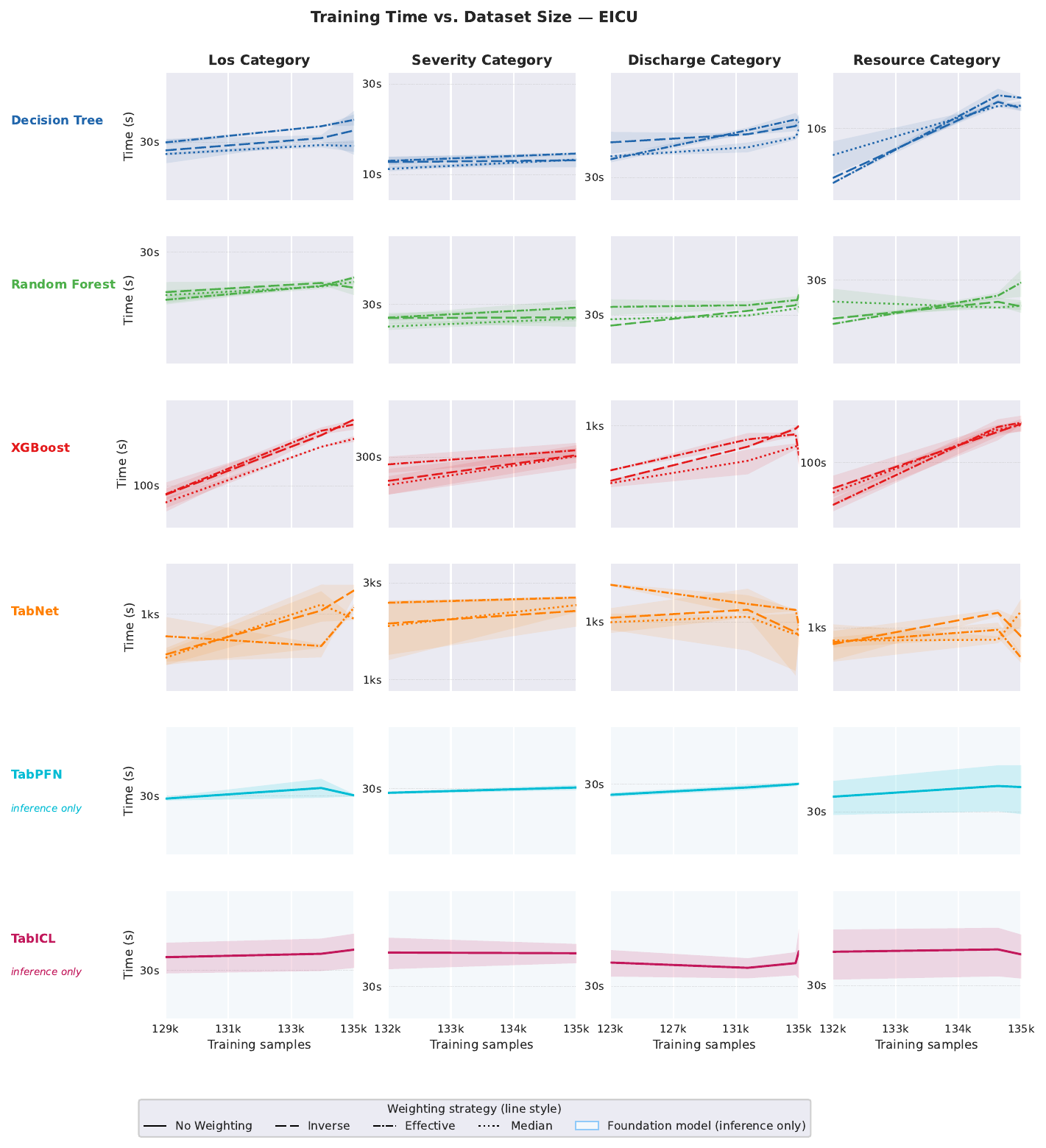}
    \caption{\textbf{Training time scaling across prediction tasks.} 
    Training time as a function of dataset size for different prediction tasks. Each panel shows the training time versus the total number of training samples for a specific target variable (length of stay, severity, discharge disposition, and resource utilization). The results are reported across all classifiers and class weighting strategies.}
    \label{fig:model_time_performance_eicu_all_targets}
\end{figure}

Figure~\ref{fig:model_time_performance_eicu_all_targets} presents extended analyses of training time scaling with dataset size across the seven prediction tasks in the eICU-CRD dataset. Training time is reported on a logarithmic scale and shown for all classifiers and weighting strategies. These plots complement the rank-based comparisons by showing absolute training durations. They suggest that training costs increase more gradually for tree-based methods than for deep learning models as sample size grows, while the inference-based foundation models remain comparatively efficient on a per-task basis.

\end{document}